\pdfoutput=1

\documentclass[11pt]{article}

\usepackage[dvipsnames]{xcolor}

\usepackage{acl}

\usepackage{times}
\usepackage{latexsym}
\usepackage{booktabs}
\usepackage{multirow}
\usepackage{array}
\usepackage{amsmath}
\usepackage{amsfonts}
\usepackage{placeins}

\usepackage[scaled=0.8]{beramono}

\usepackage{tikz}
\usetikzlibrary{shapes.geometric}
\usetikzlibrary{patterns}
\usepackage{pgfplots}
\usetikzlibrary{backgrounds}
\usetikzlibrary{matrix,calc}
\usepgfplotslibrary{fillbetween}
\pgfplotsset{compat=1.15}

\newcommand{\an}[1]{\textcolor{red}{\bf\small [#1 --AA]}}

\newcommand{\boxesplot}[9]{
    \node[rectangle,thick,draw=#3,fill=#3,opacity=0.5,text opacity=1,align=center,text width=#8,minimum height=0.5cm,anchor=south west] at (#1,1.5+#9) {#2};
    \node[rectangle,thick,draw=#5,fill=#5,opacity=0.5,text opacity=1,align=center,text width=#8,minimum height=0.5cm,anchor=south west] (a1) at (#1,0.9+#9) {#4};
    \node[rectangle,thick,draw=#7,fill=#7,opacity=0.5,text opacity=1,align=center,text width=#8,minimum height=0.5cm,anchor=south west] (a3) at (#1,0.3+#9) {#6};
    \draw[ultra thin] (#1,0+#9) -- (#1,2.1+#9);
}

\DeclareMathOperator{\bleu}{BLEU}

\usepackage[T1]{fontenc}

\usepackage[utf8]{inputenc}

\usepackage{microtype}

\title{Systematic Inequalities in Language Technology Performance \\ across the World’s Languages}

\author{Dami\'{a}n Blasi \\
  Harvard University \\
  \texttt{dblasi@fas.harvard.edu} \\\And
  Antonios Anastasopoulos \\
  George Mason University \\
  \texttt{antonis@gmu.edu} \\\And
  Graham Neubig \\
  Carnegie Mellon University \\
  \texttt{gneubig@cs.cmu.edu} \\
  }

\begin{document}
\maketitle
\begin{abstract}
Natural language processing (NLP) systems have become a central technology in communication, education, medicine, artificial intelligence, and many other domains of research and development. While the performance of NLP methods has grown enormously over the last decade, this progress has been restricted to a minuscule subset of the world's 6,500 languages. We introduce a framework for estimating the global utility of language technologies as revealed in a comprehensive snapshot of recent publications in NLP. Our analyses involve the field at large, but also more in-depth studies on both user-facing technologies (machine translation, language understanding, question answering, text-to-speech synthesis) as well as more linguistic NLP tasks (dependency parsing, morphological inflection). In the process, we (1) quantify disparities in the current state of NLP research, (2) explore some of its associated societal and academic factors, and (3) produce tailored recommendations for evidence-based policy making aimed at promoting more global and equitable language technologies.%
\footnote{
All authors contributed equally.
Data and code to reproduce the findings discussed in this paper are
available on GitHub (\url{https://github.com/neubig/globalutility}).
}
\end{abstract}

\section{Introduction}

The past decade has seen a rapid advance in natural language processing (NLP), the technology that allows computers to process human language. NLP has grown from a relatively technical niche to a fundamental tool in virtually all domains that involve language data in any shape or form. NLP is now instrumental for a vast array of tasks, from the early detection of neurodegenerative diseases \cite{orimaye2017predicting}, to exposing widespread gender and ethnic biases in societies \cite{caliskan2017semantics}, and predicting large-scale trends in collective consumer behavior \cite{kallus2014predicting}. More ostensibly, NLP has also become a staple technology for everyday frequent tasks in most contemporary societies of the world. For instance, an English speaker with a smartphone can now easily get accurate information on many topics through a quick query to a virtual assistant, they can consult an online translation service to translate a foreign language web page with a click, and they can interact with many different machines and computers through simple speech commands. 

These technological 
capabilities can be attributed to several developments over the last few decades: 1.~the advent of sophisticated machine learning methods, which allow for more effective creation of NLP systems from existing data \citep{goldberg2017neural}, 2.~the existence of standardized benchmark datasets and evaluation metrics, 3.~the prestige afforded by the research community to researchers who improve upon these benchmarks, 4.~the resulting large number of resources, be they computation, data, or ingenuity, that are poured into optimizing performance thereon. As both a theoretical and technical endeavor, NLP is experiencing an explosive increase: the annual conference of the Association of Computational Linguistics (ACL, the flagship event in NLP) received in 2000 less than 300 papers, growing in 2010 to slightly less than 1,000, to over more than 3,500 submissions in its 2020 edition. Largely as a result of this expansion of research effort, state-of-the-art systems have also achieved evaluation benchmark scores on par with human performance on a variety of NLP tasks such as question answering on English~\cite{he2021deberta}, or on automatic translation of news from German, Russian, and Chinese to English \cite{barrault-etal-2020-findings}.%
\footnote{Although the significance of these parity claims has been disputed \cite{laubli2018has}.}



These upward slanting curves on standard benchmarks fail to show how uneven this development has been for all potential NLP users. Extensive research across NLP tasks have found systematic performance drops according to dimensions such as gender, racial identity, and language varieties, among others. The reasons for these biases are multifactorial and can be traced to virtually all stages in the process of NLP development, from the data used to train systems \cite{caliskan2017semantics,sap2019risk,de2019bias,tatman-2017-gender,tatman2017effects,buolamwini2018gender,raji2019actionable}  to the very algorithms involved~\cite{speicher2018unified,bellamy2018ai,adebayo2016fairml}. The growing awareness of these biases in NLP technologies brought by these studies, along with the development of novel metrics and tests to evaluate these disparities, have resulted in progressively more efficient and principled strategies to understand and mitigate them.


However, similarly systematic approaches are still lacking in one fundamental dimension of variation across individuals: their languages. Out of the over 6,500 languages spoken or signed in the world today~\cite{hammarstrom2015ethnologue}, only a handful are systematically represented in academia and industry~\cite{joshi-etal-2020-state}. In spite of the aforementioned near-human results on translation or understanding of languages from the world's economic and political superpowers, the experience of any NLP practicioner is that, for the vast majority of languages, they fall far below such standards.
Critically, the languages of the world showcase substantial amounts of variation in most domains of description, and in fact, the performance of language technologies has been shown to be sensitive to diverse aspects of the language under study, including morphology, word order, or phonological repertoire, as well as more mundane aspects like data availability \cite{tsarfaty-etal-2020-spmrl,xia-etal-2020-predicting,arivazhagan2019massively}. Hence, the transfer of NLP developments from one language to another is far from trivial, as it often means that building highly functional language technologies on any particular language is a non-automatic, costly, and technically challenging task.

Taking all these considerations together, and given that even the consequences brought by unequal NLP technologies across (racial, gender, socioeconomic) groups within the same nominal language are already substantial, there is a pressing need for measuring and understanding NLP performance inequalities across the world's languages. Here we develop novel estimates on how the utility afforded by NLP systems is distributed across individuals, languages, and tasks at an unprecedented global scale. These estimates allow us to identify which languages are systematically under-served by language technologies and could benefit the most individuals from focused technology development. We finally trace these inequalities to the societal, economic, and academic correlates of NLP systems' performance, shedding light on its latent causes, and indicate how our results favor specific evidence-based policies in research and development.




\section{Methodology}

\begin{table*}[t]
    \centering
    \small
    \begin{tabular}{@{}p{.28\textwidth}|p{.4\textwidth}|p{.2\textwidth}@{}}
    \toprule
         \textbf{Task} & \textbf{Description} & \textbf{Metric} \\
         \midrule
         Syntactic Analysis (DEP) & Infer syntactic dependencies between words in text & Labeled Attachment Score \\[.3em]
         Morphological Inflection (ING) & Produce an inflection given a lemma and morphological tags & Accuracy \\[.3em]
         Machine Translation (MT) & Translate text from a language into another & BLEU score \\ [.3em]
         Speech Synthesis (TTS) & Produce speech on the basis of textual input & 1-mel-cepstral distortion \\[.3em]
         Natural Language Inference (NLI) & Recognize entailment or contradiction between two sentences & Accuracy \\[.3em]
         Question Answering (QA) & Produce an answer for a textual query  & Fscore \\
         \bottomrule
    \end{tabular}
    \caption{NLP tasks evaluated in the present study, along with their corresponding performance metric.}
    \label{tab:tasks}
\end{table*}

\subsection{Quantifying utility and demand}
Our fundamental goal is evaluating the distribution of diverse representative language technologies (and their qualities) across the world's languages and their populations. Minimally, we would attempt to account for the patterns of association between the {\em demand} of language technologies and the {\em utility} they confer to users across languages. Thus, the first component of our analysis pertains quantifying the {\em utility} users in a given language $l$ receive from a language technology. Ideally, such a measure would capture to what extent a given NLP system solves the specific problems an individual can pose to them - for instance, how successful an automatic translation is in translating a webpage, or how faithfully a speech recognition system is in executing a series of verbal commands.  Intuitively, utility is associated with the nominal performance of the technology - in NLP systems more specifically, performance is typically measured by contrasting the solution offered by the machine against the one a (knowledgeable) human would provide. How this comparison is instantiated and measured depends on the task (see Section \ref{tab:tasks}); however, since our purpose is to allow for comparisons, we define the utility of a task and language, $u_l$, as the corresponding performance normalized by the best possible performance afforded by such task, i.e.
\begin{align*}
u_{l}=\frac{\textrm{performance}_l}{\textrm{theoretical max performance}}
\end{align*}

In cases where the best possible performance is undefined or technically unattainable, we take the empirical maximum as an estimate of the theoretical one and normalize by the best-performing language across all languages $L$, i.e. we replace the denominator in the above definition by $\max_{l' \in L}(\textrm{performance}_{l'})$.

Defining utility in this manner allow us to explore and contrast language technologies at the broadest scale, which is possible thanks to some necessary simplifying assumptions. As we pointed out before, not all users of the same language technology might benefit in the same manner given a fixed utility, and the relation between nominal performance and ``true" utility might be complex and non-linear.

With these caveats in mind, we further quantify the second component of our analysis, the {\em demand} for a language technology in each language $l$, $d_l$. We characterize $d_{l}$ by taking into consideration demographic and linguistic perspectives. Under the first perspective, the demand for a given technology in a language is estimated to be proportional to the number of speakers of the language itself $n_{l}$  ($d_{l} \propto n_{l}$). Under the second perspective, the demand across the approximately 6,500 languages of the world is identical ($d_{l} \propto 1$). These two alternatives as well as any intermediate combination of them can be simply parameterized through a single exponent $\tau$,
\begin{align*}
   d_{l}^{(\tau)} = \frac{n_{l}^\tau}{\sum_{l' \in  L} n_{l'}^\tau}
\end{align*}
where $\tau=1$ correspond to a demographic notion of demand, $\tau=0$ to a linguistic one, and $0<\tau<1$ is in between.

Equipped with these notions, we construct a simple family of global  metrics ($M_\tau$) revealing to what degree the global demand for language technologies is actually met:
\begin{align*}
   M_{\tau} = &\sum_{l \in \mathcal{L}} d_{l}^{(\tau)} \cdot u_{l}
\end{align*}

$M_{\tau}$ has a number of intuitive properties we would like such a metric to have. $M_\tau$ is bounded between 0 and 1; 0 corresponds to a case where no-one benefits from a given language technology, whereas 1 would correspond to a situation where all languages enjoy perfect technology. Increasing the utility of a given language leads to an increase in $M_{\tau}$, and the magnitude of this increase is influenced by both the size of the improvement and the demand in that language.

\subsection{NLP tasks}
We apply our measures of utility and demand to a set of diverse and major representative NLP tasks, which are described below and summarized in Table~\ref{tab:tasks}.

The first three are tasks that technology users interact with directly in their everyday life, so that their output is already in a shape and form that is usable for most individuals. \emph{Question answering} (QA) consists of crafting a relevant answer to a question formulated in natural language, such as e.g. ``what is the capital city of the Philippines?" or ``why do dogs like bones?". This task is ubiquitous in online search or virtual assistants. \emph{Machine translation} (MT) is the task of translating from one language to another (e.g. from Tagalog to Estonian or from Japanese to Basque), and is typically used to facilitate inter-personal communication, information gathering, and e-commerce.
\emph{Text-to-speech} (TTS) is the task of rendering speech from textual input, which is used widely in spoken virtual assistants, car navigation systems, and in general is becoming the standard gateway for the internet of things.

Beyond these three user-facing tasks, we also consider three more technical and linguistically-focused tasks, which often inform part of the pipelines of the user-facing tasks but which are rarely if ever encountered ``in the wild" by language technology users.\emph{Morphological Inflection} (Inflection) is the task of generating an inflected wordform given a lemma and a morphological specification, e.g. producing the third person singular form for ``run'': \texttt{run+3;SG}$\rightarrow$\texttt{runs}. 
\emph{Syntactic Parsing} under the dependency formalism (DEP) is the task of producing a syntactic parse of an input sentence, e.g. given the sentence ``dogs like bones'' specifying the ``dogs'' and ``bones'' are the subject and object of ``like'' respectively. \emph{Natural Language Inference} (NLI) is a central task in AI and involves the evaluation of information presented in propostional format. More specificially, given a sentence called the ``premise'' (e.g.~``the dog chewed a big bone''), NLI systems decide whether a separate sentence called the ``hypothesis'' is entailed by the premise (e.g.~``the dog gnawed at a bone''), negated by it (e.g.~``the dog was sleeping''), or neither (e.g.~``the dog likes bones'').

\begin{figure*}[t]
    \centering
    \includegraphics[width=1\linewidth]{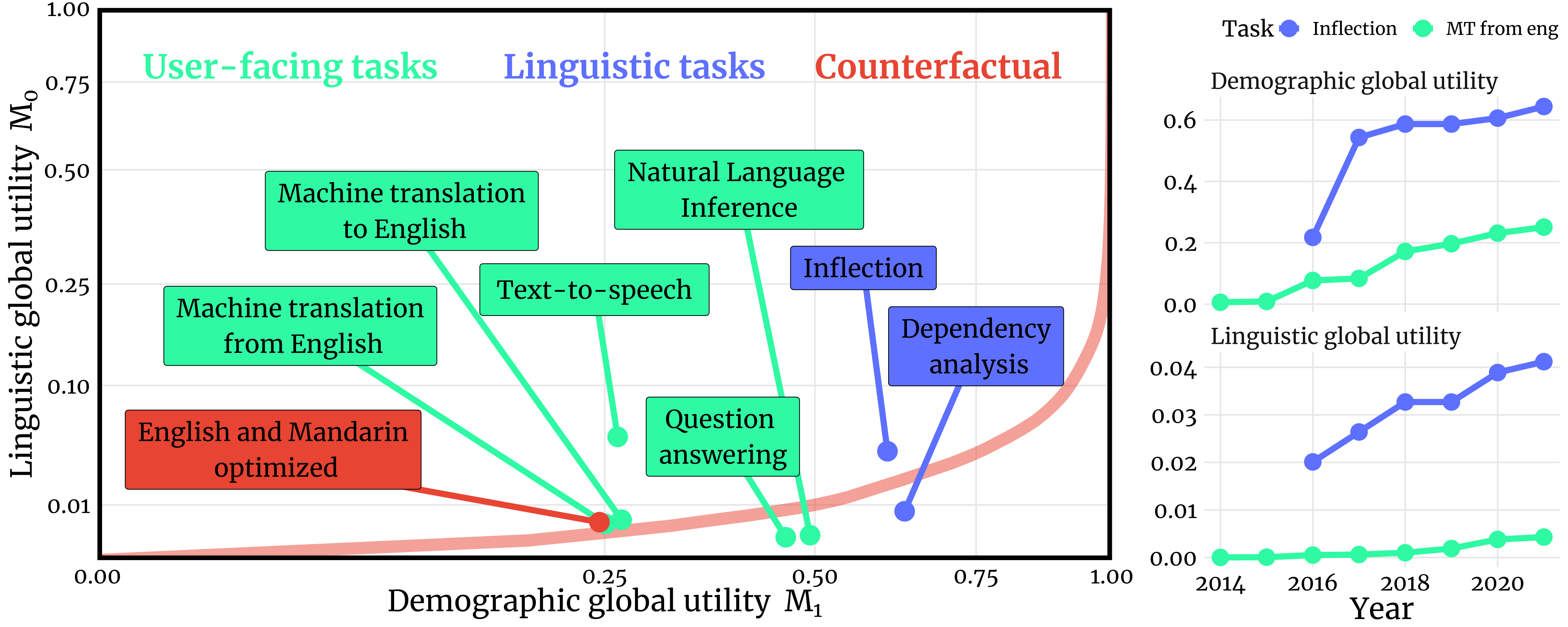}
    \caption{Left panel: linguistic and demographic global utility metrics for a number of language technology tasks. The red curve corresponds to the sequence where first the  language with the largest number of users is set to utility 1, then the second, and so on. Right panel: recent historical progression of two language technology tasks: Inflection and Machine Translation from English.}
    \label{fig:main}
\end{figure*}

\subsection{Correlates of NLP utility}
Beyond the performance of individual tasks, we take a bird's-eye-view of the field of language technologies in general, as we analyze some of the correlates of the scientific production in NLP. In particular, we follow two broad guiding questions: (1) does the system of academic incentives promote the development of a more linguistically diverse NLP? and (2) is economic centrality or sheer demographic demand the best predictor of NLP technologies in any given language?

While a full understanding of the complex causal mechanisms binding society and NLP in general is outside of the scope of the present article, we set out to provide a first large-scale exploration of these matters by considering scientific publications appearing in major international NLP conferences as the basic units of science production. This simplification is not without challenges: for instance, some widely used language technologies are developed outside of the traditional scientific circuit based on proprietary technology, or they are published in local conferences, possibly in languages other than English.\footnote{e.g.~the Japanese NLP society's 2020 conference published 396 papers: \url{https://www.anlp.jp/proceedings/annual_meeting/2020/}} In spite of this, studying scientific publications (and their correlates) allows us to evaluate transparent questions on the basis of publicly available data at a scale that is unfeasible for in-depth analyses.

Therefore, we study the first question by determining whether the cumulative number of citations a paper receives is correlated with the number of languages it is associated with. We investigate our second question by finding the best predictive model of the number of NLP papers in any given language by contrasting two predictors: estimated number of users worldwide and approximate GDP associated with its users. We model these regression problems in a Bayesian generalized mixed effects framework (see \autoref{sec:appendix_methods}).

\subsection{Data}

We manually collect information on task performance for a number of diverse representative NLP technologies, as summarized in Table \ref{tab:tasks} (see Materials \& Methods in  \autoref{sec:appendix_materials}). These range from user-facing applications like {\em machine translation} (i.e.~the automatic translation of text in one language into another) to more linguistic NLP tasks such as {\em dependency parsing} (i.e.~the analysis of syntactic or semantic relationships between words). The data is taken from a combination of multilingual benchmarks, shared tasks and published results in NLP conferences. Demographic and linguistic information necessary for the estimation of demands were obtained from a variety of sources, including Ethnologue, Glottolog, and the World Trade Organisation.

\begin{figure*}
    \centering
    \begin{tabular}{c@{}c}
        Dependency Parsing: $M_1=0.63$ & Morphological Inflection: $M_1=0.64$ \\
        \includegraphics[width=.48\textwidth]{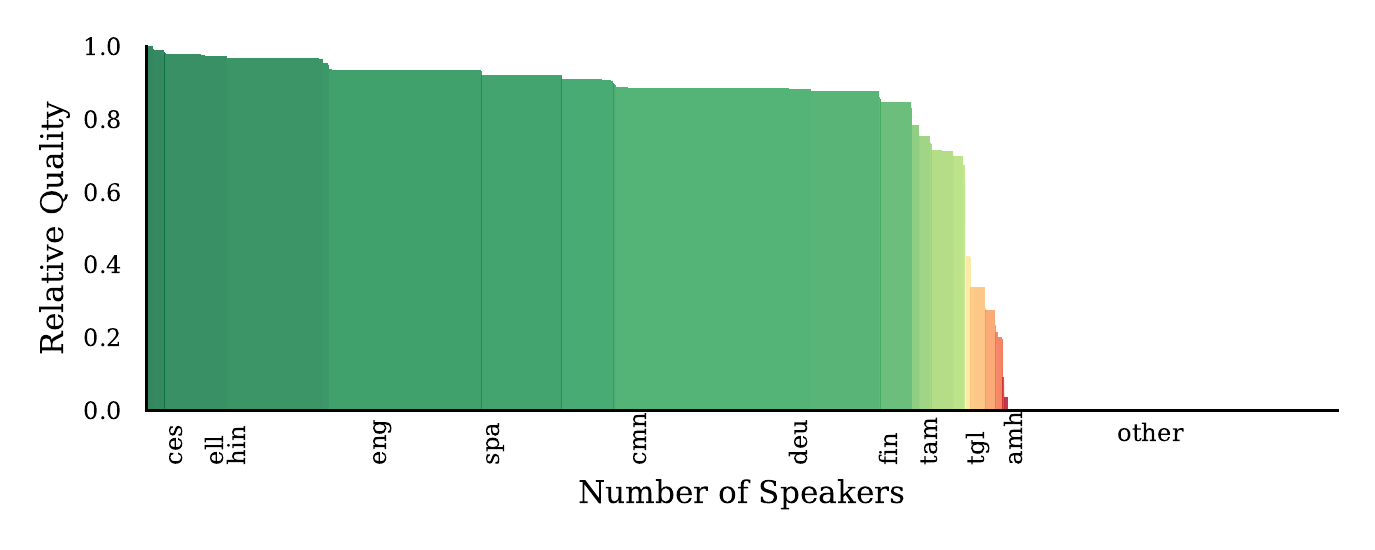}  &  \includegraphics[width=.48\textwidth]{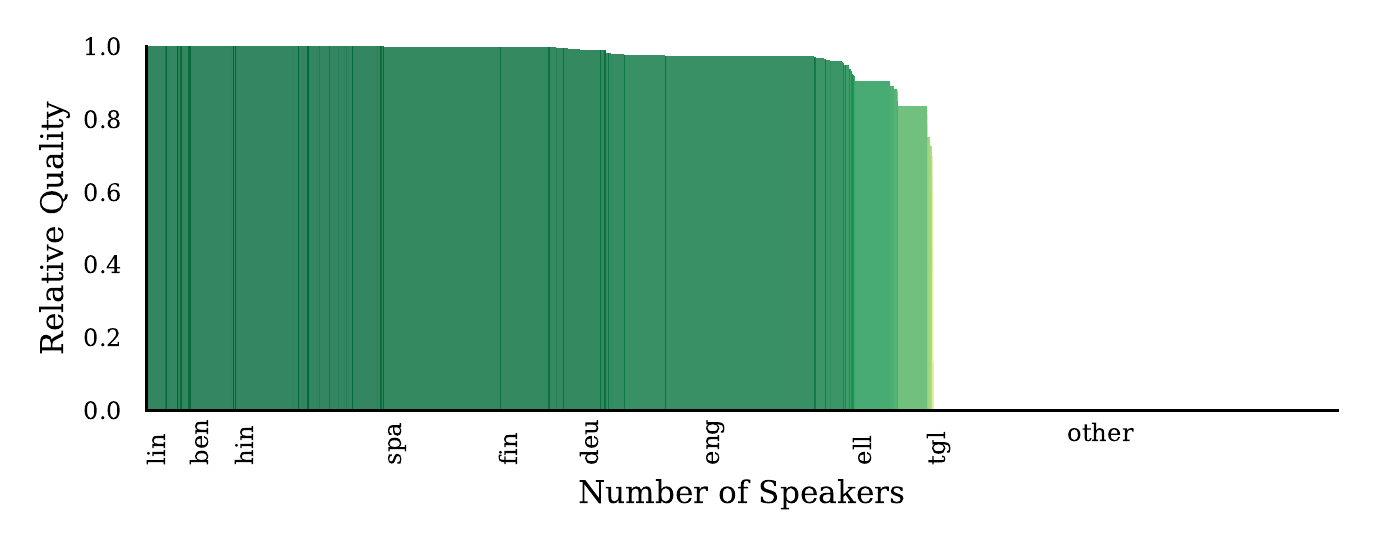} \\
        Natural Language Inference: $M_1=0.42$ & Question Answering: $M_1=0.36$ \\
        \includegraphics[width=.48\textwidth]{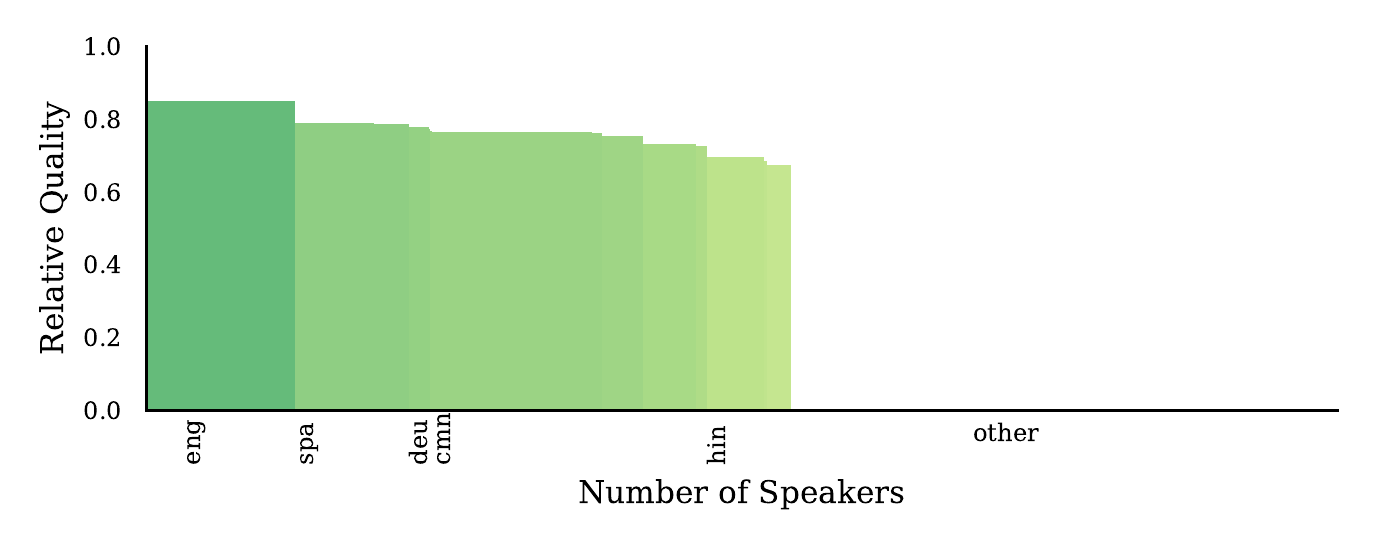}  &  \includegraphics[width=.48\textwidth]{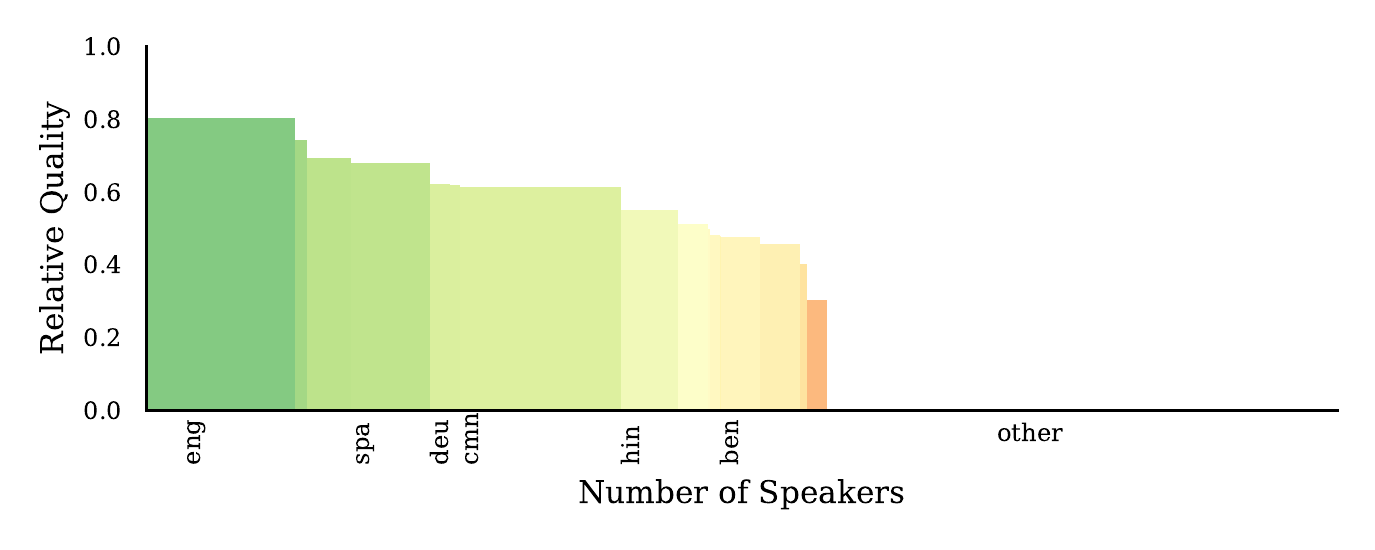} \\
        Speech Synthesis: $M_1=0.32$ & Machine Translation (X$\rightarrow$English): $M_1=0.49$ \\
        \includegraphics[width=.48\textwidth]{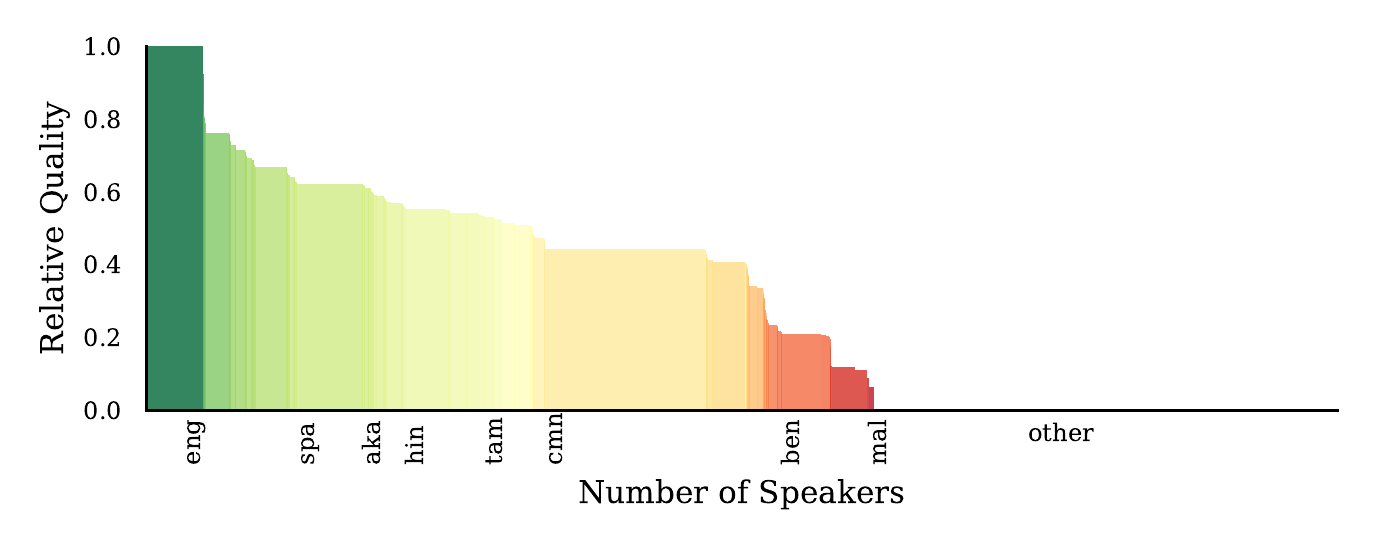}  &  \includegraphics[width=.48\textwidth]{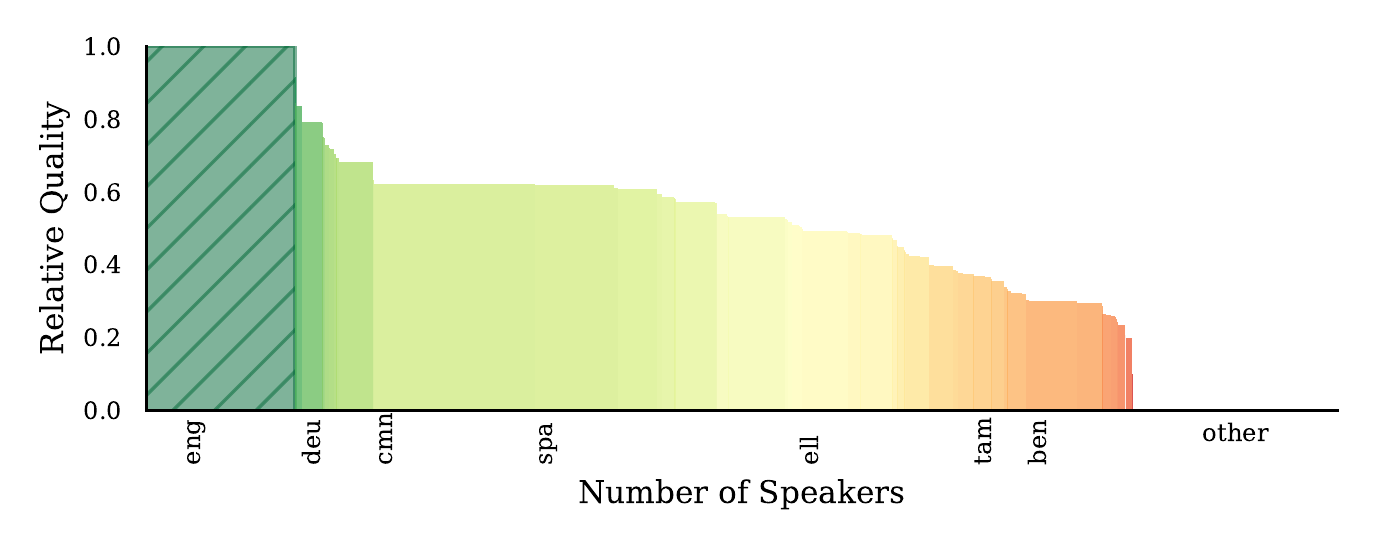} \\
        Machine Translation (X$\rightarrow$Spanish): $M_1=0.36$ & Machine Translation (X$\rightarrow$Bengali): $M_1=0.10$ \\
        \includegraphics[width=.48\textwidth]{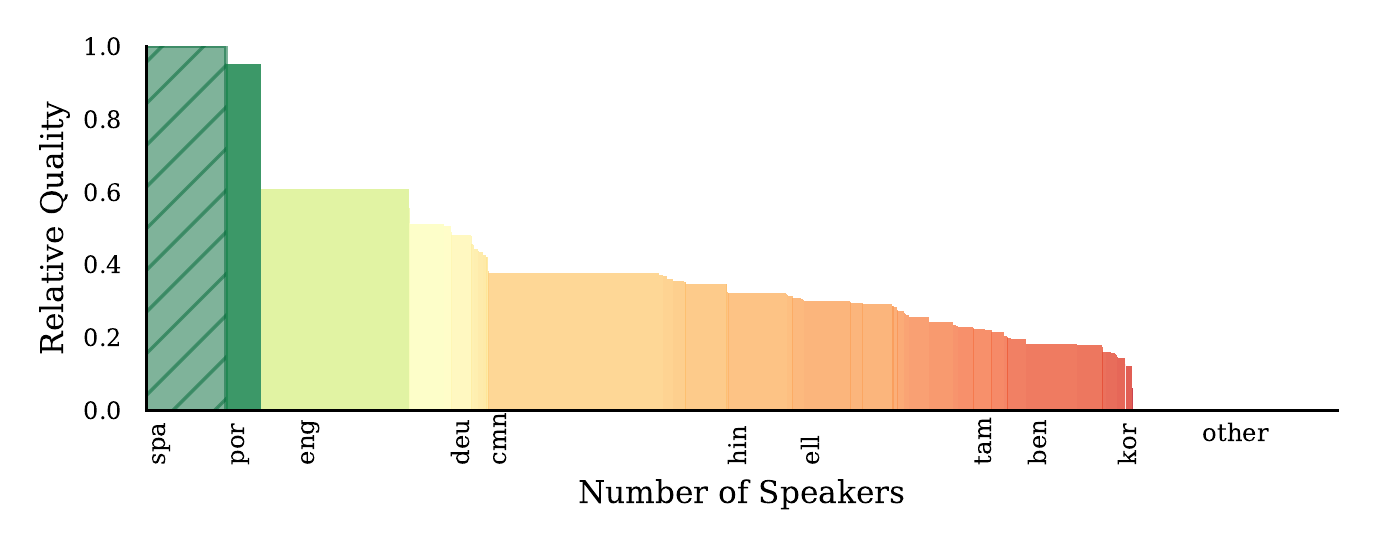}  &  \includegraphics[width=.48\textwidth]{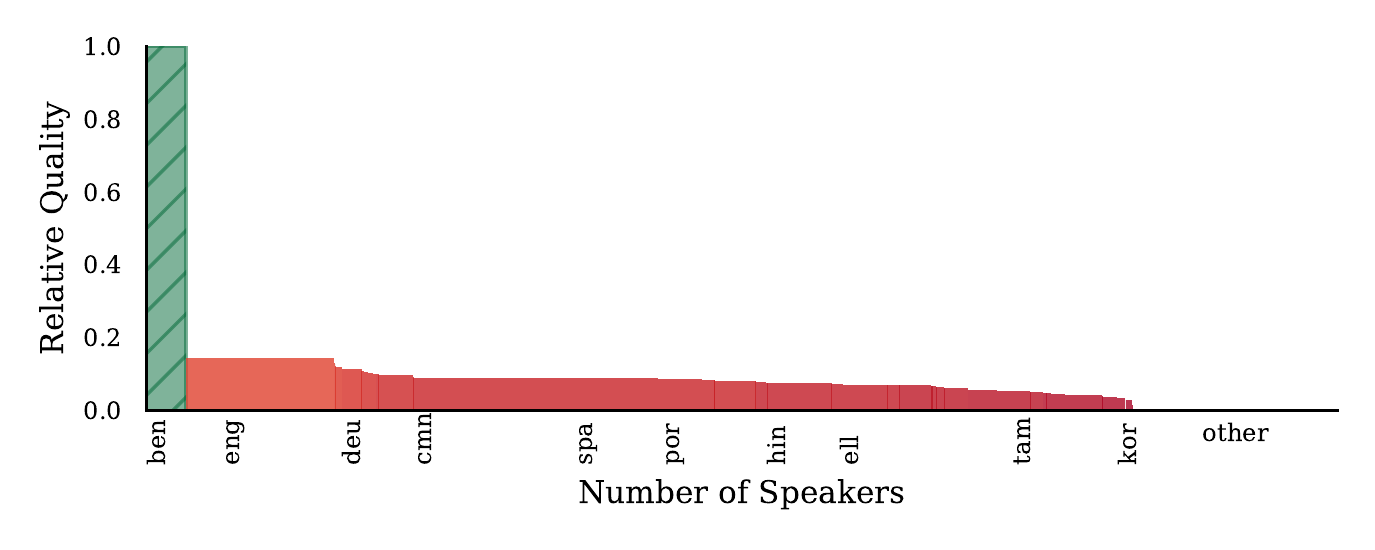}\\
        QA [on Arabic Vernaculars]: $M_1^{\text{ara}}=0.58$ & QA  [on Swahili Vernaculars]: $M_1^{\text{swa}}=0.23$\\
        \includegraphics[width=.48\textwidth]{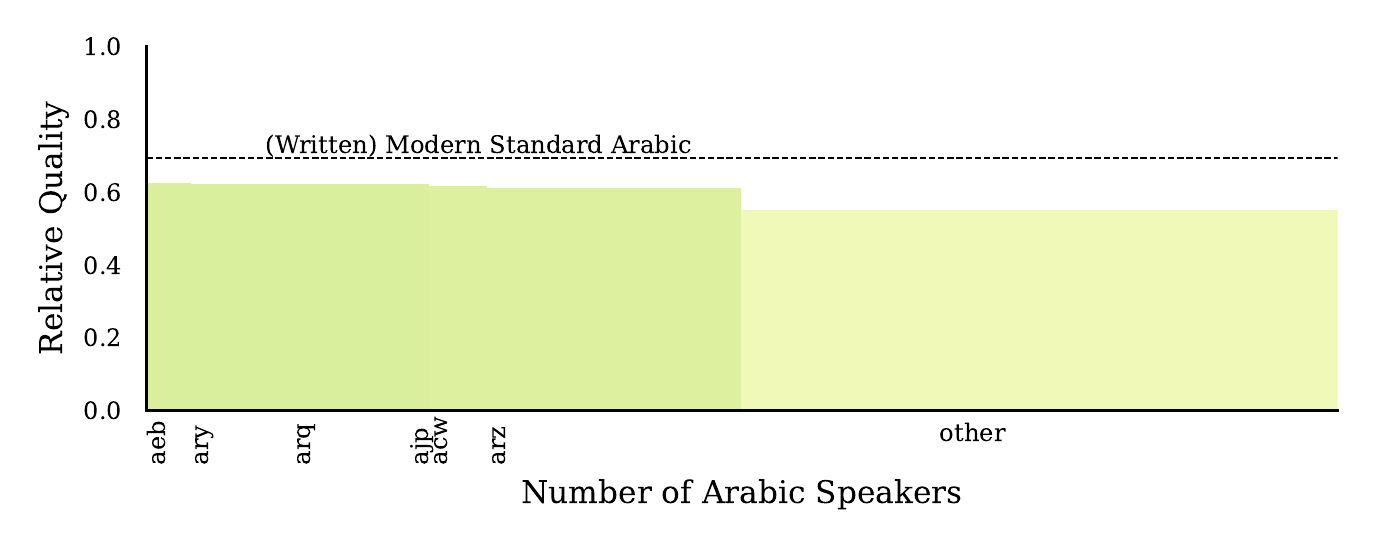}  &  \includegraphics[width=.48\textwidth]{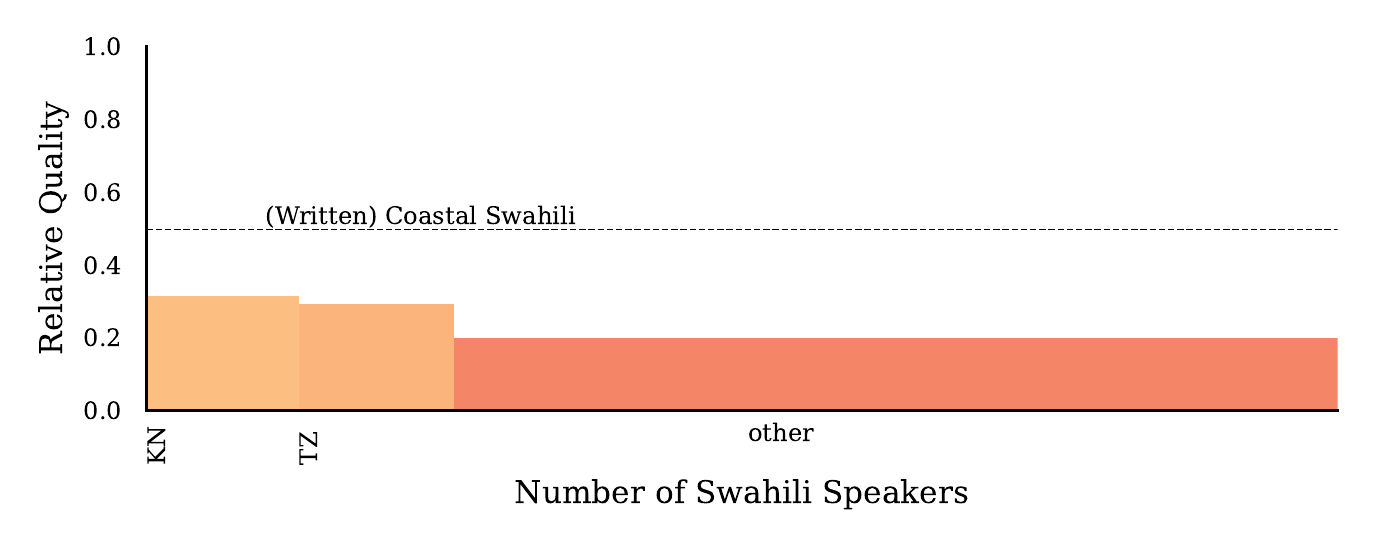}\\
        \multicolumn{2}{l}{\footnotesize{acw: Hijazi Arabic, aeb: Tunisian Arabic,  ajp: South Levantine Arabic, aka: Aka, amh: Amharic, arq: Algerian Arabic,}} \\
        \multicolumn{2}{l}{\footnotesize{ary: Moroccan Arabic, arz: Egyptian Arabic, ben: Bengali, ces: Czech, cmn: Mandarin Chinese, deu: High German,}}\\
        \multicolumn{2}{l}{\footnotesize{ell: Greek, eng: English, fin: Finnish, hin: Hindi, kor: Korean, lin: Lingala, mal: Malayalam, por: Portuguese,}}\\
        \multicolumn{2}{l}{\footnotesize{spa: Spanish, swa: Swahili, tam: Tamil, tgl: Tagalog.}}
    \end{tabular}
    \caption{Illustration of our metric on demographic-focused utility ($\tau=1$) on various NLP tasks.}
    \label{fig:metrics}
\end{figure*}

\section{Results and Analysis} 

\subsection{General observations}
Figure~\ref{fig:main} presents an overview of our main findings. Unsurprisingly, most NLP tasks we focus on fare substantially better when utility is measured demographically rather than linguistically. 

Text-to-speech synthesis is the task with the most linguistic coverage: the published results (due to a single study~\cite{black2019cmu}) cover more than 630 languages (or about 10\% of the world's languages). However, for the vast majority of these languages the measured quality of the generated speech is about half as good as the exceptionally good English system~\cite{ren2020fastspeech}. The next most linguistically diverse tasks are those regarding morphosyntactic analysis, i.e. morphological inflection and dependency parsing, 
which have been evaluated over 140 and 90 languages respectively. 
For these more esoteric tasks which do not necessarily convey direct utility to a downstream user, the majority of the systems are in general very good.

Natural language inference (NLI; a representative natural language understanding task) 
and question answering (QA) lie on the opposite side of the spectrum: the established benchmarks have only focused on up to 15 and 17 languages respectively, leading to very low scores on the linguistic axis.  

In Figure~\ref{fig:main} (right panel) we observe the progress of the utility metrics in  tasks for which we had access to comparable data across a span of the last 7 years. The extensive efforts of the UniMorph project \cite{kirov2018unimorph} to cover as many languages as possible are visible in the ``Inflection'' plot, with significant improvements over time. On the other hand, the machine translation field is still in the process of ramping up following demographics and/or socioeconomic priorities, with improved linguistic coverage over the years.

The granularity of these findings can be increased on the basis of available data.
Figure~\ref{fig:metrics} additionally presents demographic utility across language populations for all tasks. The visualization allows for identification of ostensive gaps in received utility. 
The two bottom plots of Figure~\ref{fig:metrics} display our metrics over speakers of a single language, based on question answering results for different spoken Arabic and Swahili lectal varieties~\cite{faisal-etal-21-sdqa}. This analysis shows that utility differences are small between Arabic vernaculars although these systems still lag behind the systems for Modern Standard Arabic, while the utility level of Coastal Swahili speakers in Tanzania is about 10\% lower than that for speakers in Kenya. 

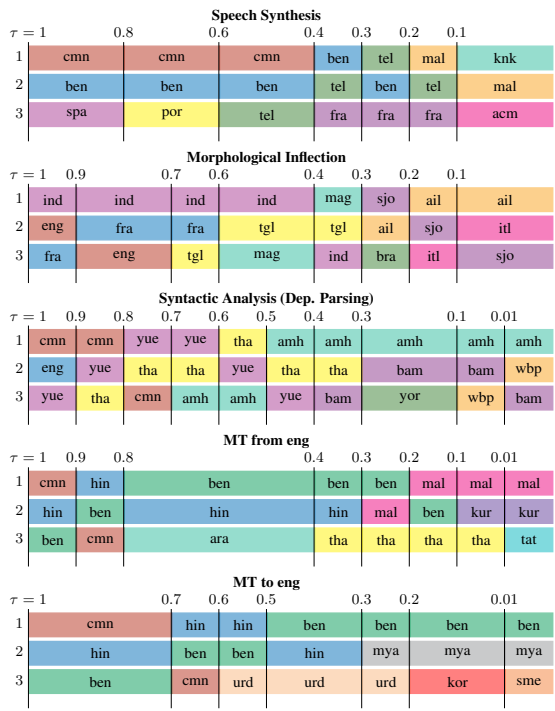
\begin{figure}
    \centering
    \def\MTY{0}
    \def\MTE{3}
    \def\DEP{6}
    \def\INF{9}
    \def\TTS{12}
    \small
    \resizebox{.45\textwidth}{!}{%
    \begin{tikzpicture}[every node/.style={inner sep=0pt}]
        \node[anchor=south] at (5,2.5+\MTY) {\textbf{MT to eng}};
        \node[anchor=east] at (-0.1,1.8+\MTY) {\small $1$};
        \node[anchor=east] at (-0.1,1.2+\MTY) {\small $2$};
        \node[anchor=east] at (-0.1,0.6+\MTY) {\small $3$};
        \boxesplot{0}{cmn}{BrickRed}{hin}{NavyBlue}{ben}{ForestGreen}{3cm}{\MTY};
        \node[anchor=south] at (0,2.2+\MTY) {\small $\tau=1$};
        \boxesplot{3}{hin}{NavyBlue}{ben}{ForestGreen}{cmn}{BrickRed}{1cm}{\MTY};
        \node[anchor=south] at (3,2.2+\MTY) {\small $0.7$};
        \boxesplot{4}{hin}{NavyBlue}{ben}{ForestGreen}{urd}{Apricot}{1cm}{\MTY};
        \node[anchor=south]  at (4,2.2+\MTY) {\small $0.6$};
        \boxesplot{5}{ben}{ForestGreen}{hin}{NavyBlue}{urd}{Apricot}{2cm}{\MTY};
        \node[anchor=south]  at (5,2.2+\MTY) {\small $0.5$};
        \boxesplot{7}{ben}{ForestGreen}{mya}{Gray}{urd}{Apricot}{1cm}{\MTY};
        \node[anchor=south]  at (7,2.2+\MTY) {\small $0.3$};
        \boxesplot{8}{ben}{ForestGreen}{mya}{Gray}{kor}{red}{2cm}{\MTY};
        \node[anchor=south]  at (8,2.2+\MTY) {\small $0.2$};
        \boxesplot{10}{ben}{ForestGreen}{mya}{Gray}{sme}{Orange}{1cm}{\MTY};
        \node[anchor=south]  at (10,2.2+\MTY) {\small $0.01$};
        
        \node[anchor=south] at (5,2.5+\MTE) {\textbf{MT from eng}};
        \node[anchor=south] at (0,2.2+\MTE) {\small $\tau=1$};
        \node[anchor=east] at (-0.1,1.8+\MTE) {\small $1$};
        \node[anchor=east] at (-0.1,1.2+\MTE) {\small $2$};
        \node[anchor=east] at (-0.1,0.6+\MTE) {\small $3$};
        \boxesplot{0}{cmn}{BrickRed}{hin}{NavyBlue}{ben}{ForestGreen}{1cm}{\MTE};
        \node[anchor=south] at (1,2.2+\MTE) {\small $0.9$};
        \boxesplot{1}{hin}{NavyBlue}{ben}{ForestGreen}{cmn}{BrickRed}{1cm}{\MTE};
        \node[anchor=south] at (2,2.2+\MTE) {\small $0.8$};
        \boxesplot{2}{ben}{ForestGreen}{hin}{NavyBlue}{ara}{SeaGreen}{4cm}{\MTE};
        \node[anchor=south]  at (6,2.2+\MTE) {\small $0.4$};
        \boxesplot{6}{ben}{ForestGreen}{hin}{NavyBlue}{tha}{yellow}{1cm}{\MTE};
        \node[anchor=south]  at (7,2.2+\MTE) {\small $0.3$};
        \boxesplot{7}{ben}{ForestGreen}{mal}{RubineRed}{tha}{yellow}{1cm}{\MTE};
        \node[anchor=south]  at (8,2.2+\MTE) {\small $0.2$};
        \boxesplot{8}{mal}{RubineRed}{ben}{ForestGreen}{tha}{yellow}{1cm}{\MTE};
        \node[anchor=south]  at (9,2.2+\MTE) {\small $0.1$};
        \boxesplot{9}{mal}{RubineRed}{kur}{RoyalPurple}{tha}{yellow}{1cm}{\MTE};
        \node[anchor=south]  at (10,2.2+\MTE) {\small $0.01$};
        \boxesplot{10}{mal}{RubineRed}{kur}{RoyalPurple}{tat}{Aquamarine}{1cm}{\MTE};
        
        \node[anchor=south] at (5,2.5+\DEP) {\textbf{Syntactic Analysis (Dep. Parsing)}};
        \node[anchor=south] at (0,2.2+\DEP) {\small $\tau=1$};
        \node[anchor=east] at (-0.1,1.8+\DEP) {\small $1$};
        \node[anchor=east] at (-0.1,1.2+\DEP) {\small $2$};
        \node[anchor=east] at (-0.1,0.6+\DEP) {\small $3$};
        \boxesplot{0}{cmn}{BrickRed}{eng}{RoyalBlue}{yue}{Mulberry}{1cm}{\DEP};
        \node[anchor=south] at (1,2.2+\DEP) {\small $0.9$};
        \boxesplot{1}{cmn}{BrickRed}{yue}{Mulberry}{tha}{yellow}{1cm}{\DEP};
        \node[anchor=south] at (2,2.2+\DEP) {\small $0.8$};
        \boxesplot{2}{yue}{Mulberry}{tha}{yellow}{cmn}{BrickRed}{1cm}{\DEP};
        \node[anchor=south]  at (3,2.2+\DEP) {\small $0.7$};
        \boxesplot{3}{yue}{Mulberry}{tha}{yellow}{amh}{SeaGreen}{1cm}{\DEP};
        \node[anchor=south]  at (4,2.2+\DEP) {\small $0.6$};
        \boxesplot{4}{tha}{yellow}{yue}{Mulberry}{amh}{SeaGreen}{1cm}{\DEP};
        \node[anchor=south]  at (5,2.2+\DEP) {\small $0.5$};
        \boxesplot{5}{amh}{SeaGreen}{tha}{yellow}{yue}{Mulberry}{1cm}{\DEP};
        \node[anchor=south]  at (6,2.2+\DEP) {\small $0.4$};
        \boxesplot{6}{amh}{SeaGreen}{tha}{yellow}{bam}{Fuchsia}{1cm}{\DEP};
        \node[anchor=south]  at (7,2.2+\DEP) {\small $0.3$};
        \boxesplot{7}{amh}{SeaGreen}{bam}{Fuchsia}{yor}{OliveGreen}{2cm}{\DEP};
        \node[anchor=south]  at (9,2.2+\DEP) {\small $0.1$};
        \boxesplot{9}{amh}{SeaGreen}{bam}{Fuchsia}{wbp}{YellowOrange}{1cm}{\DEP};
        \node[anchor=south]  at (10,2.2+\DEP) {\small $0.01$};
        \boxesplot{10}{amh}{SeaGreen}{wbp}{YellowOrange}{bam}{Fuchsia}{1cm}{\DEP};
        
        \node[anchor=south] at (5,2.5+\INF) {\textbf{Morphological Inflection}};
        \node[anchor=south] at (0,2.2+\INF) {\small $\tau=1$};
        \node[anchor=east] at (-0.1,1.8+\INF) {\small $1$};
        \node[anchor=east] at (-0.1,1.2+\INF) {\small $2$};
        \node[anchor=east] at (-0.1,0.6+\INF) {\small $3$};
        \boxesplot{0}{ind}{Mulberry}{eng}{BrickRed}{fra}{RoyalBlue}{1cm}{\INF};
        \node[anchor=south] at (1,2.2+\INF) {\small $0.9$};
        \boxesplot{1}{ind}{Mulberry}{fra}{RoyalBlue}{eng}{BrickRed}{2cm}{\INF};
        \node[anchor=south]  at (3,2.2+\INF) {\small $0.7$};
        \boxesplot{3}{ind}{Mulberry}{fra}{RoyalBlue}{tgl}{yellow}{1cm}{\INF};
        \node[anchor=south]  at (4,2.2+\INF) {\small $0.6$};
        \boxesplot{4}{ind}{Mulberry}{tgl}{yellow}{mag}{SeaGreen}{2cm}{\INF};
        \node[anchor=south]  at (6,2.2+\INF) {\small $0.4$};
        \boxesplot{6}{mag}{SeaGreen}{tgl}{yellow}{ind}{Mulberry}{1cm}{\INF};
        \node[anchor=south]  at (7,2.2+\INF) {\small $0.3$};
        \boxesplot{7}{sjo}{Fuchsia}{ail}{YellowOrange}{bra}{OliveGreen}{1cm}{\INF};
        \node[anchor=south]  at (8,2.2+\INF) {\small $0.2$};
        \boxesplot{8}{ail}{YellowOrange}{sjo}{Fuchsia}{itl}{RubineRed}{1cm}{\INF};
        \node[anchor=south]  at (9,2.2+\INF) {\small $0.1$};
        \boxesplot{9}{ail}{YellowOrange}{itl}{RubineRed}{sjo}{Fuchsia}{2cm}{\INF};

        \node[anchor=south] at (5,2.5+\TTS) {\textbf{Speech Synthesis}};
        \node[anchor=south] at (0,2.2+\TTS) {\small $\tau=1$};
        \node[anchor=east] at (-0.1,1.8+\TTS) {\small $1$};
        \node[anchor=east] at (-0.1,1.2+\TTS) {\small $2$};
        \node[anchor=east] at (-0.1,0.6+\TTS) {\small $3$};
        \boxesplot{0}{cmn}{BrickRed}{ben}{RoyalBlue}{spa}{Mulberry}{2cm}{\TTS};
        \node[anchor=south] at (2,2.2+\TTS) {\small $0.8$};
        \boxesplot{2}{cmn}{BrickRed}{ben}{RoyalBlue}{por}{yellow}{2cm}{\TTS};
        \node[anchor=south]  at (4,2.2+\TTS) {\small $0.6$};
        \boxesplot{4}{cmn}{BrickRed}{ben}{RoyalBlue}{tel}{OliveGreen}{2cm}{\TTS};
        \node[anchor=south]  at (6,2.2+\TTS) {\small $0.4$};
        \boxesplot{6}{ben}{RoyalBlue}{tel}{OliveGreen}{fra}{Fuchsia}{1cm}{\TTS};
        \node[anchor=south]  at (7,2.2+\TTS) {\small $0.3$};
        \boxesplot{7}{tel}{OliveGreen}{ben}{RoyalBlue}{fra}{Fuchsia}{1cm}{\TTS};
        \node[anchor=south]  at (8,2.2+\TTS) {\small $0.2$};
        \boxesplot{8}{mal}{YellowOrange}{tel}{OliveGreen}{fra}{Fuchsia}{1cm}{\TTS};
        \node[anchor=south]  at (9,2.2+\TTS) {\small $0.1$};
        \boxesplot{9}{knk}{SeaGreen}{mal}{YellowOrange}{acm}{RubineRed}{2cm}{\TTS};
    \end{tikzpicture}
    }
    \caption{The priority languages (top-3 shown) change with different balancing of demographic and linguistic utility, with focus shifting from populous languages e.g. Mandarin (cmn) and Hindi (hin) to more under-served languages.}
    \label{fig:rankings}
\end{figure}

\subsection{Priorities in NLP development}
Given the current snapshot of NLP systems, we could ask which languages will lead to the largest global utility improvement. The relative importance of linguistic vs. demographic demands determines the priority ranking, as it can be observed in Figure~\ref{fig:rankings} for a sample of five tasks. 
Improving on the demographic-focused utility entails a greater emphasis on Mandarin Chinese, Hindi, Spanish, and other populous languages that are generally well-served by current technologies. Balancing linguistic and demographic considerations leads to prioritizing a more diverse set of languages, mostly Asian and African languages like Amharic, Bambara, Bengali, Thai, or Yoruba, which are both populous and under-served, along with also large but severely under-served languages like Kurdish, Urdu, and Oromo. Further emphasis on linguistic utility would lead to prioritization of indigenous and potentially endangered languages of small communities like Aimele, Itelmen, North Sami, or Warlpiri, which are currently largely ignored by NLP research~\cite{bird-2020-decolonising}.

\begin{figure*}[t!]
    \centering
    \includegraphics[width=1\linewidth]{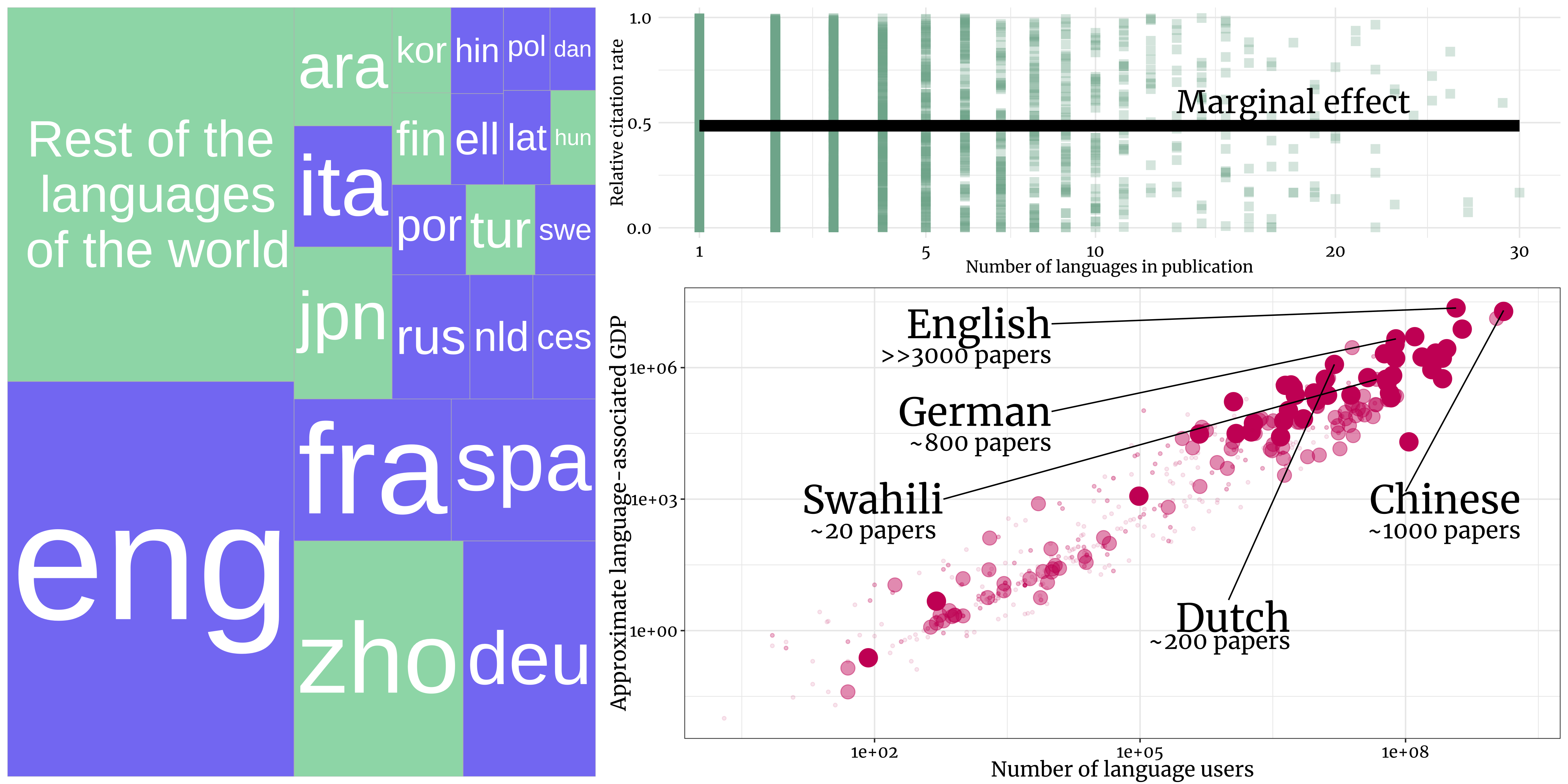}
    \caption{Left panel: treemap of the number of NLP publications per language (with area  proportional to the number). eng: English, zho: Chinese, deu: German, fra: French, spa: Spanish, jpn: Japanese, rus: Russian, nld: Dutch, ces: Czech, por: Portuguese, tur: Turkish, swe: Swedish, ita: Italian, fin: Finnish, ell: Greek, lat: Latin, hun: 
    Hungarian, ara: Arabic, kor: Korean, hin: Hindi, pol: Polish, dan: Danish. Right top panel: Relative citation rate vs number of languages in the publication. Right bottom panel: Number of publications according to number of language users and approximate GDP. Point size and transparency scales with number of publications.}
    \label{fig:society}
\end{figure*}

\subsection{The role of society, economy, and academia}
Now we turn to our large-scale analysis of NLP publications.
First, this reveals that a substantial proportion of publications do not even describe in a clear and unequivocal manner the language (or languages) they are dealing with~\cite{bender2011achieving}. Given the current prevalence of English of a language of study in NLP, in most cases, the lack of an explicit reference to a particular language entails the system deals with English exclusively.

This perhaps reflects a more deep-seated issue at play reflected in the citation of papers over time. Independently of publication venue, year, or subfield of NLP research, the number of languages a publication deals with is not predictive of how many citations it will accrue over time (see Figure~\ref{fig:society}, top right panel). In other words, if citations can be regarded as a proxy for academic incentives, scientists and developers are presented with little to no additional academic reward when tackling data, problems, or tasks involving more than one language.

This naturally leads to the question of what explains the production of language technologies across languages to start with, which will necessarily involve agents, mechanisms, and data, outside of the scope of NLP publications themselves. Nevertheless, in order to contribute to this investigation, we determined whether approximate measures of economic centrality or number of language users were better predictors of sheer number of papers published for any given language (see \autoref{sec:bibliometric}). While both variables are substantially collinear, we find that approximate GDP (rather than number of users) leads to a substantially smaller prediction error of number of published papers.

\section{Discussion}

Our study, covering diverse NLP tasks and types of evidence, makes apparent the immense inequality in the development of language technologies across the world's languages. After English, a handful of Western European languages dominate the field -in particular German, French, and and Spanish- as well as even fewer non-Indo-European languages, primarily Chinese, Japanese, and Arabic. Our preliminary investigation suggests it is the economic prowess of the users of a language (rather than the sheer demographic demand) what drives the development of language technologies.

In spite of this, for some tasks (such as Inflection) there is an encouraging trend of both demographic- and linguistic-utility improving year-over-year. This is due to the nature of the task; reasonably accurate solutions can be achieved through small but highly-curated data. Since linguistic expertise on the languages of the world is, naturally, globally distributed, the main hurdle these tasks face is to pool such expertise under the premise of a common technical goal. In this respect, relatively low-cost and bottom-up actions that gather experts to work on specific NLP tasks (such as Universal Dependencies and UniMorph) have succeeded in accelerating the cross-linguistic development of language technologies. These prosper mainly on the basis of academic incentives, as those individuals or groups who contribute data and/or expertise  are rewarded with individual publications or co-authorship in collective publications. Many of these contributions - which do not necessarily involve hefty resource investments but instead linguistic expertise - are markedly different from the typical publications in language technologies.

However, these more esoteric tasks are tenuously associated with those that users are more likely to interact with, such as Machine Translation or Speech Synthesis. User-facing tasks all have in common a tight dependency on computational resources and large data, which in turn hinge on substantial financial means. In a context of pressing user needs across multiple populations and languages, we submit that future developments on policies aimed at furthering cross-linguistic technologies would benefit from clear (and possibly standardized) metrics that assist in streamlining complex decisions regarding resource allocation. Our measures of global coverage fulfill that role, and help identifying large but currently under-served languages. While we do not attempt to supplement the necessary in-depth evaluation of the need of each individual group and language, they provide a common ground for coordinating global efforts across heterogeneous actors.

\section*{Acknowledgements}
This work was supported by NSF Award 2040926.

\bibliography{References}
\bibliographystyle{acl_natbib}

\appendix

\section{Materials}
\label{sec:appendix_materials}

\paragraph{Publication data}

We rely on papers available through the Anthology of the Association of Computational Linguistics\footnote{\url{https://www.aclweb.org/anthology/}} which hosts more than 60 thousand papers from all major NLP conferences. We rely on Semantic Scholar~\cite{ammar-etal-2018-construction} for citation information.

We make the working assumption that a mention of a language in a research paper likely entails that the underlying research involves this language. We follow an automatic pipeline for finding language mentions in a paper, which starts by converting the paper PDF to a machine-readable format. We then search within the paper for any mention of a language's English name(s), its endonym, as well as its ISO or Glottolog code. We then apply a post-processing step to ensure the precision of this pipeline as our simple text-based search is prone to false positives for languages whose names match common English words (e.g. She, Male, Label, Even, The, Are), common placenames (e.g. Colorado, Nara, Sydney), parts of author names (e.g.  Su, Kim, Dan, Ali, Rama), or mathematical notation (e.g. Dji, Dii).

In addition, we enrich each publication by imputing its research area. There were 16 research areas identified, based on the ones represented at recent major NLP conferences (specifically starting with the 2019 version of EMNLP, and removing some of the areas that were unique to that conference). For each area, we identified 1-6 publication venues from the ACL Anthology, where more venues were chosen when each venue had relatively few publications. Based on the abstracts of papers from each of these venues, we trained a bag-of-words classifier using the linear support vector machine implementation in scikit-learn\footnote{\url{https://scikit-learn.org/stable/}}, and applied this classifier to the abstracts of the papers we wanted to classify. Necessary data and code to reproduce these results are released in the supplementary material.

\paragraph{Data Sources and Metrics for Utility}

The majority of NLP research relies on automatic evaluation metrics over datasets annotated with gold-standard outputs.
The advantage of this approach is that it allows consistent comparisons between systems and a seamless evaluation of progress on a specific evaluation set.
On the other hand, there is no guarantee that even statistically significant improvement on an automatic metric translates to improvements on user-perceived utility.
Nevertheless, the reality is that virtually all published NLP research reports automatic evaluation metrics, with only a tiny fraction diverging from the norm by e.g. using human evaluations.

Our analysis assumes that all named languages have standard versions that are comprehensible and acceptable to all members of the population identified as ``speakers” in our sources. However, we have the demographic information necessary for more fine-grained analysis in only a handful of languages. While this assumption is certainly an oversimplification, we nevertheless believe it does not detract from our paper's arguments. 

For a completely fair comparison across languages, one would ideally compute automatic metrics over the same or an equally representative evaluation set. For our language understanding case study this requirement is satisfied, as the XNLI~15 language test sets are translations of the same evaluation set. Utility in this case, where the evaluation metric $m$ is accuracy, will be equal to the accuracy for each language's $l$ test set: $\mathrm{utility}(l,m) = m_l$. 

Natural language understanding results are sourced from the XNLI leaderboard~\cite{conneau2018xnli}, which contains test datasets with premise-hypothesis pairs in 15 languages. 

For question answering (QA) we aggregate results from two established multilingual benchmarks, namely TyDi-QA~\cite{clark2020tydi} and MLQA~\cite{lewis2009ethnologue}. Both benchmarks focus on extractive question answering, i.e. finding the text span of a given document that answers, if possible, a given question. The two benchmarks jointly cover 17 languages. We keep the highest results for languages that are shared between the two datasets (English and Arabic). For this task we equate utility with test set F-score, a measure that meaningfully combines precision and recall of the retrieved answer span.

For machine translation, we collected more than~500 published MT results from all WMT and IWSLT evaluation campaigns, as well as more than 50 MT studies from the last three years' ACL, EMNLP, and NAACL conferences~\citep{acl-2017-association,acl-2018-association,emnlp-2017-2017,emnlp-2018-2018,naacl-2016-2016,naacl-2018-2018,acl-2019-association,emnlp-2019-2019,emnlp-2020-2020}.  In the machine translation field the most popular evaluation metric is BLEU \cite{papineni2002bleu}. In our MT case studies we estimate utility based on a normalized version of BLEU, such that for translation from $s$ to $t$ with $\bleu(s,t)$ over an established test set, we have $\mathrm{utility}(s,t,\bleu) \approx \frac{\bleu(s,t)}{Z}$.
The normalizing factor $Z = \max_{\mathcal{L}\times\mathcal{L}} \bleu$ is equivalent to the largest reported BLEU, which we equate to the largest attainable utility at the snapshot of interest. In all our MT case studies we use $Z=70$, which is the BLEU score reported for translation between Serbian and Croatian \cite{arcan2016asistent}. 

For text-to-speech synthesis, we relied on results from the CMU Wilderness project \cite{black2019cmu}, which builds TTS voices with FestVox~\cite{anumanchipalli2011festvox}, and compared them to the English system of~\cite{ren2020fastspeech}. The quality of the synthesized audio is evaluated using mel-cepstral distortion \cite[MCD]{kubichek1993mel} a distortion measure that compares synthesized examples with originals (lower is better). Each MCD of $x_l$ for a language $l$ was converted to a relative utility score by applying the transformation $\frac{x_{\max}-x_l}{x_{\max} - x_{\min}}$, where $x_{\max}$ and $x_{\min}$ correspond to the highest (worst) and lowest (best) observed MCD scores across all languages.

For syntactic analysis through dependency parsing, we relied on results from two state-of-the-art systems, UDPipe~\cite{straka2018udpipe} and UDify~\cite{kondratyuk201975}. The systems are typically evaluated using two measures, Unlabeled and Labeled Attachment Score (UAS and LAS), which measure the overlap between human-created and automatically-produced syntactic trees, excluding punctuation. For our metrics we use LAS, which considers the semantic relation (e.g. Subj) used to label the attachment between two words.

The results on morphological inflection were taken from the findings of the corresponding shared tasks that have been taking place as part of the SIGMORPHON workshop for the past 5 years~\cite{cotterell-etal-2016-sigmorphon,cotterell-etal-2017-conll,cotterell-etal-2018-conll,mccarthy-etal-2019-sigmorphon,vylomova-etal-2020-sigmorphon}. 
The systems are evaluated using exact-match accuracy over a pre-defined test set in each language, simply comparing the correct inflected form with the system's output.

\paragraph{Population Demand}
We compile population statistics from various sources. We rely on Ethnologue \cite{ethnologue2018} for language population statistics.  We take special care when computing population statistics over macro-languages (e.g. Arabic, Chinese) and languages commonly spoken by L2 speakers (e.g. English) or across multiple dialects (e.g. for Spanish or Portuguese), aggregating populations across all variants.

\paragraph{Economic Indicators for Demand}

We aggregate economic information on international trade, as provided from the World Trade Organisation (WTO) through the World Integrated Trade Solution.\footnote{\url{https://wits.worldbank.org/}} Since each language community can be geographically associated with a member nation of WTO, we can then estimate economic indicators for and between language communities.\footnote{Our conclusions and analyses based on WITS data are the responsibility of the authors and do not represent the opinion of the WTO.}

In a monolingual setting, we rely on the most recent GDP estimates, associated with each language community. For example, the 1.7 million Nahuatl speakers represent about 1.3\% of Mexico's population, and thus the final GDP associated with the Nahuatl language will be 1.3\% of Mexico's GDP.

Modeling demand in a bilingual setting (across two languages) is also feasible using economic indicators. For instance, the amount of trade between two language communities could be used to approximate the need for translation between the two.
Specifically, if we use the normalized import volume per language community then we can estimate demand for an $s\rightarrow t$ translation system as
$\mathrm{demand}(s,t) \propto v^{\text{import}}_{s\rightarrow t}$
such that $\sum_{s \in \mathcal{L}} v^{\text{import}}_{s\rightarrow t} = 1$.

Take the Azerbaijani language as an example: Azerbaijan's imports mainly come from the Russian Federation (16.8\%), Turkey (14.7\%), China (11.2\%), the US (8.5\%), Ukraine (5.5\%), and Germany (5.5\%).\footnote{Source: \url{https://wits.worldbank.org/CountryProfile/en/Country/AZE/Year/2017/TradeFlow/Import}} Hence, we can assign a proportional weight to model demand for translation from Russian, Turkish, Chinese, English, Ukrainian, and German into Azerbaijani respectively. One could equivalently use the normalized volume of exports instead.

This is only straightforward to compute in cases where a language is easy to map to a specific country. In cases of languages that are commonly used across many countries e.g. German (which is the main language in both Germany and Austria) or macro-languages spoken in larger regions of the world, we combine the weights accordingly in order to jointly model the demand for the whole language community.

Table~\ref{tab:demand_econ} presents the top-15 translation pairs based on demand estimated from economic indicators, namely the import (and export) partner share of the target (source) language. We note that this ranking does not take underlying populations into account, using only the \textit{percentage} of demand for each language community. Several entries in Table~\ref{tab:demand_econ} are language pairs that are rarely, if ever, studied in MT case studies, like Belarusian-Russian, Mongolian-Mandarin Chinese, Albanian-Italian, or Russian-Armenian.

\section{Methods}
\label{sec:appendix_methods}

\paragraph{Predicting Utility on Unseen Languages/Pairs}
\label{sec:estimating}

One of the main disadvantages of using solely published results for estimating quality and, hence, utility, is the lack of evaluations on all languages or language pairs. Furthermore, not all languages or pairs are consistently evaluated on newly developed models.
To counter this issue, we propose a more comprehensive approach which attempts to predict the expected quality/utility over languages or language pairs unseen in the collected literature.

A naive approach is to make the approximation that utility on any unseen language is~0. However crude, this could be a valid assumption in many cases: consider the example of a language understanding system trained on all languages that appear in Wikipedia. Such a system, without proper modifications, would not be able to handle input in Yupik or Dhivehi (Maldivian), since these languages are not represented in Wikipedia and they use different writing systems than any other language.
Note that, in such a case, for a language understanding system evaluated over a classification task as in a language understanding setting, the expected utility is not~0, but is rather the expected quality of random outputs (33\% in the case of three-way classification).

\paragraph{Estimating MT quality with pivoting}

In the case of machine translation, pivoting is a viable approach for producing translations between any arbitrary language pair, as long as the intermediate systems exist.
Even if no published results exist on translation from German to Chinese, it is unreasonable to assign an expected utility of~0 to such a MT system, since there exist high-quality German-English and English-Chinese systems.

In the case of cascaded systems, though, estimating utility requires a careful approach, due to error propagation. Consider a system A with accuracy~80\% and a system B with accuracy also~80\%. A cascaded system where the output of system A is provided as input to system B will have an expected accuracy~64\%, not~80\%.

An important point is that there is no reason for pivoting through a single language. Consider the example of Catalan to Chinese translation. A path from Catalan to Spanish, to English, to Chinese might have a yield a higher estimated utility from a single-language pivoting path, since its components are of higher quality.

We devise a method that allows us to generalize this notion in order to find the highest estimated utility for every language pair.
We construct a weighted directed graph $\mathcal{G}$=$(V,E)$ with each node $v \in V$ representing a language. 
The weighted directed edge $e_{s\rightarrow t}$ between nodes $s$ and $t$ will have a weight equal to the highest reported normalized BLEU score on translation from $s$ to $t$. If no results have been published on this language pair, we set the weight of that edge to~0.

With graph $\mathcal{G}$ in hand, as long as a path from nodes $s$ to $t$ exists, we can estimate the expected normalized BLEU of $s-t$ translation as the maximum cumulative (multiplicative) weight over any path from $s$ to $t$. If a path does not exist, then the estimation is 0. This is possible in cases where a language is reported as only source or only target in the literature; for example, Greek (ell) only appears as a source in a single study (reporting Greek--English translation results) which allows us to estimate Greek--X utility by pivoting through English, but we cannot produce estimates for X--Greek. Table~\ref{tab:pivoting} presents translation pairs were our estimated utility (normalized BLEU score) is higher than the published results.

\section{Bibliometric Analysis}
\label{sec:bibliometric}

\paragraph{Analysis of Citations}
To each publication we associate its citation percentile relative to its year and event. 
We analyze normalized citations ($C$) through Bayesian generalized additive mixed effects models implemented in R with brms and Stan \cite{burkner2017brms,carpenter2017stan} We utilize default weakly informative priors for all parameters and we run four MCMC chains  for each model which in all cases achieved convergence. The distribution of $C$ is described through a beta distribution, of which its expected value is given by
\begin{equation}
    \mathbb{E}[C]=\textrm{logit}(f(L) + \alpha_{A} + \beta_{A}\cdot L)
\end{equation}
where $f(L)$ is a smooth function (on the basis of thin plate splines) depending on the number of languages dealt with in the paper ($L$), and $\alpha_{A}$ and $\beta_{A}$ are random intercepts and slopes according to each area, respectively. In order to evaluate the support in favor of $f(L)$, we compared the leave-one-out (LOO) performance of this model against a counterpart without this term,
\begin{equation}
    \mathbb{E}[C]=\textrm{logit}(\alpha_{A} + \beta_{A}\cdot L)
\end{equation}
The difference in expected log pointwise predictive density (which serves to inform model selection, \citep{vehtari2017practical})  between the two models is -0.9 (SE=0.6), which implies there is no major performance difference between the two.

\paragraph{Analysis of Number of Publications}
We determine the total estimated number of papers in which each language $l$ was involved ($P_{l}$). The resulting distribution has a large concentration of zero values, so we opt to model this through a zero-inflated negative binomial distribution. We focus on two parameters: the expected value of the number of publications ($\mathbb{E}[P]$) and the mixture probability ($\pi$). In both cases, we fit models considering three possibilities: (1) A smooth (thin plate spline) function of the log-GDP, (2) a smooth (thin plate spline) function of the log-number of speakers, and (3) a fixed parameter. This leads to evaluating 9 models through a LOO criterion. The model that involves (1) for both parameters displays the best overall performance (see SI).

\section{Machine Translation Case Studies}
\label{sec:mt_casestudy}

We use this section to expand on the discussion of MT case studies.

\paragraph{Translation involving English} 
Since translation involves two languages and language communities, there are two natural ways for a speaker to receive utility from a MT system: either by being the \textit{source} (with their language being translated into another) or by having another language translated into theirs (\textit{target}). We disentangle the two by only using each one at a time for our utility calculations.

Utilities based on demographics for both settings are similar, with $M=0.25$ (from English) and $M_1=0.27$ (to English). Since published results only cover 101 languages, the linguistic diversity scores are much lower, with $M_0$ around $0.005$. 


\paragraph{Translation among all languages}
We extend our study on translation among all languages (still maintaining the distinction between a language used as source or target). We base our estimates for utility on any reported results, as well as on accuracy estimates based on a pivoting approach. Briefly outlined, our pivoting estimation approach finds the best performing translation path for language pairs without reported results, i.e. since no studies report translation accuracy when translating from Greek to Chinese, we find that among all possible translation paths, translating from Greek to English and from English to Chinese yields the highest expected accuracy. We outline the process in the Materials and Methods section.

\begin{figure*}
    \centering
    \begin{tabular}{@{}c@{}c@{}c@{}}
    ara  $\rightarrow  X$ & aze $\rightarrow  X$ & ben $\rightarrow  X$ \\
    \includegraphics[width=.32\linewidth]{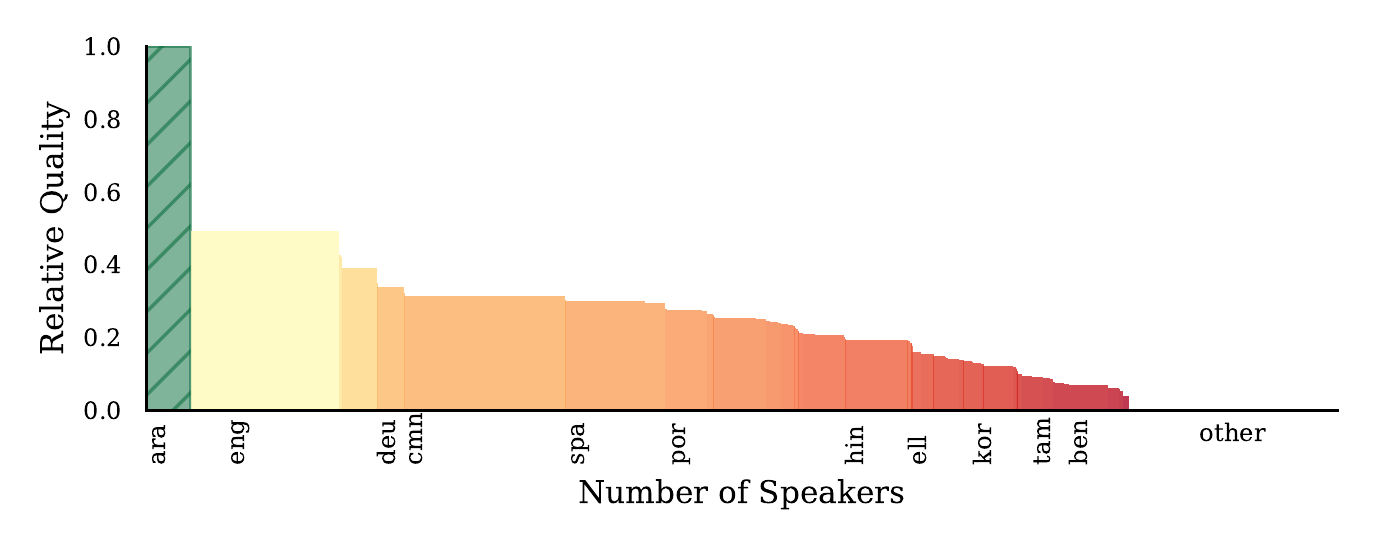} & \includegraphics[width=.32\linewidth]{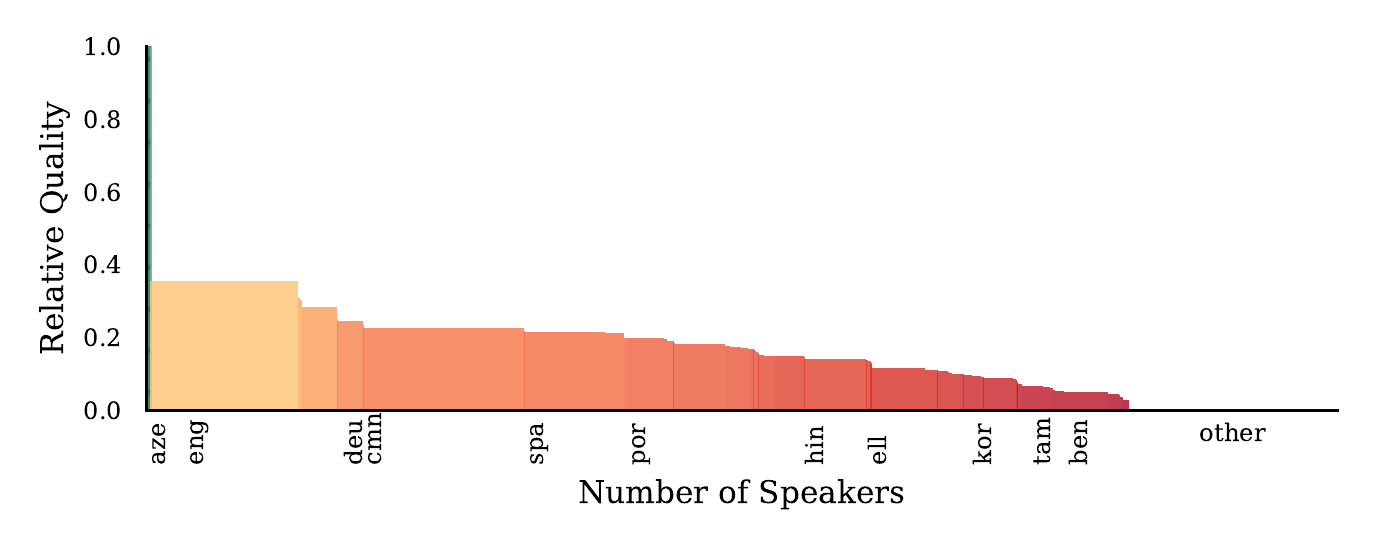} & \includegraphics[width=.32\linewidth]{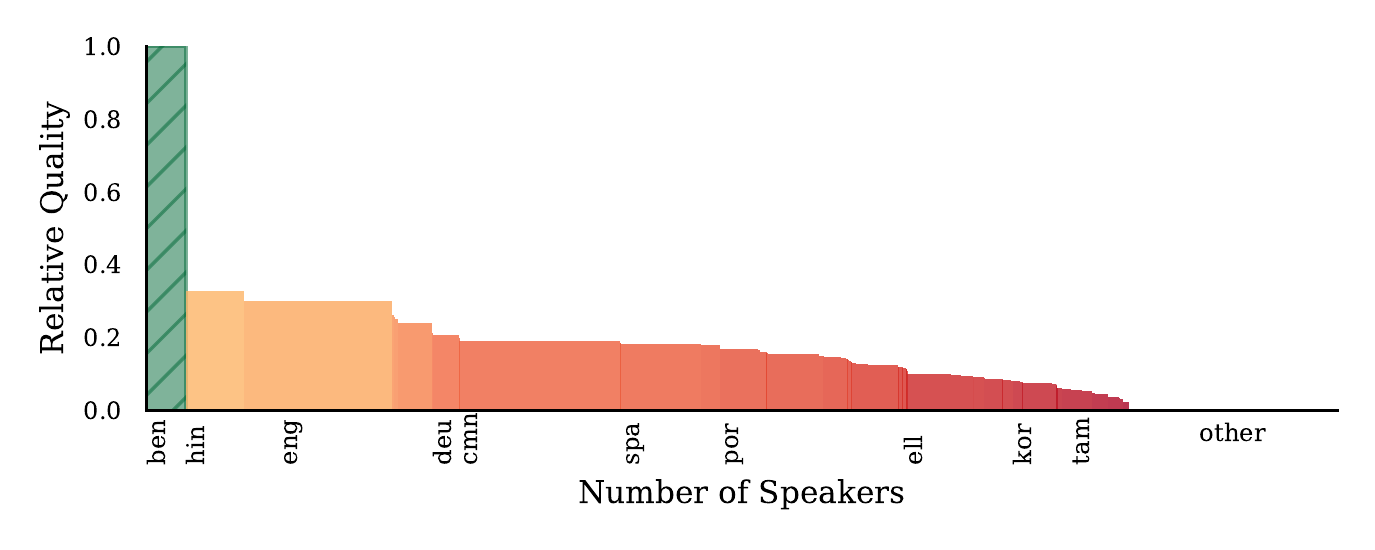} \\
    cat $\rightarrow  X$ & cmn $\rightarrow  X$ & deu $\rightarrow  X$ \\
    \includegraphics[width=.32\linewidth]{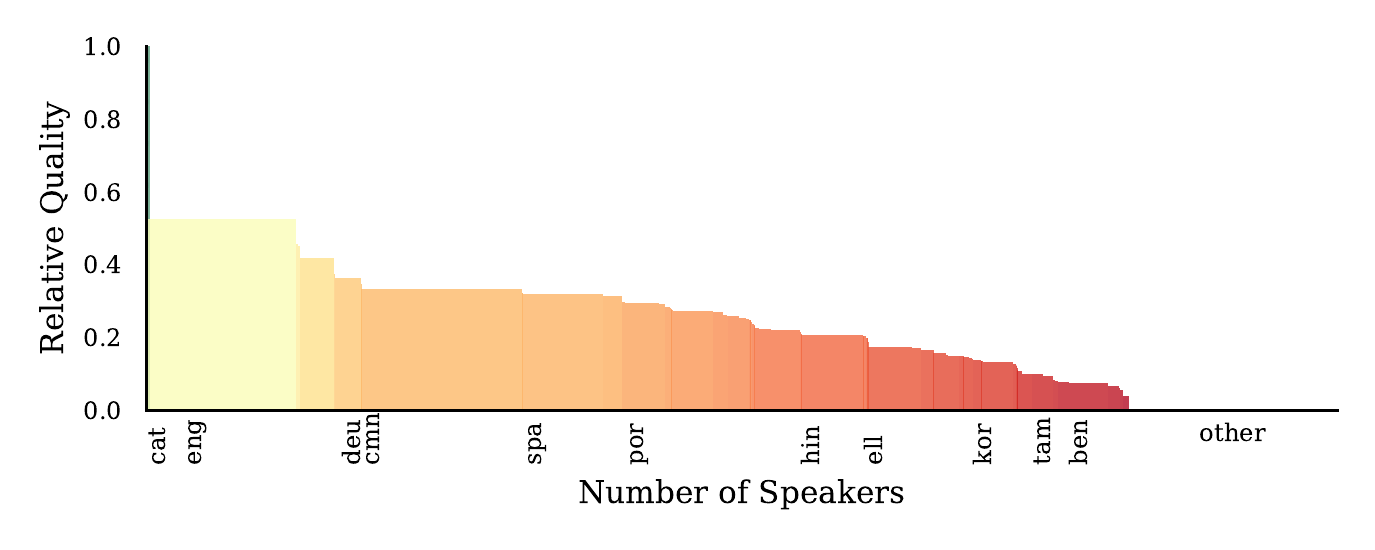} & \includegraphics[width=.32\linewidth]{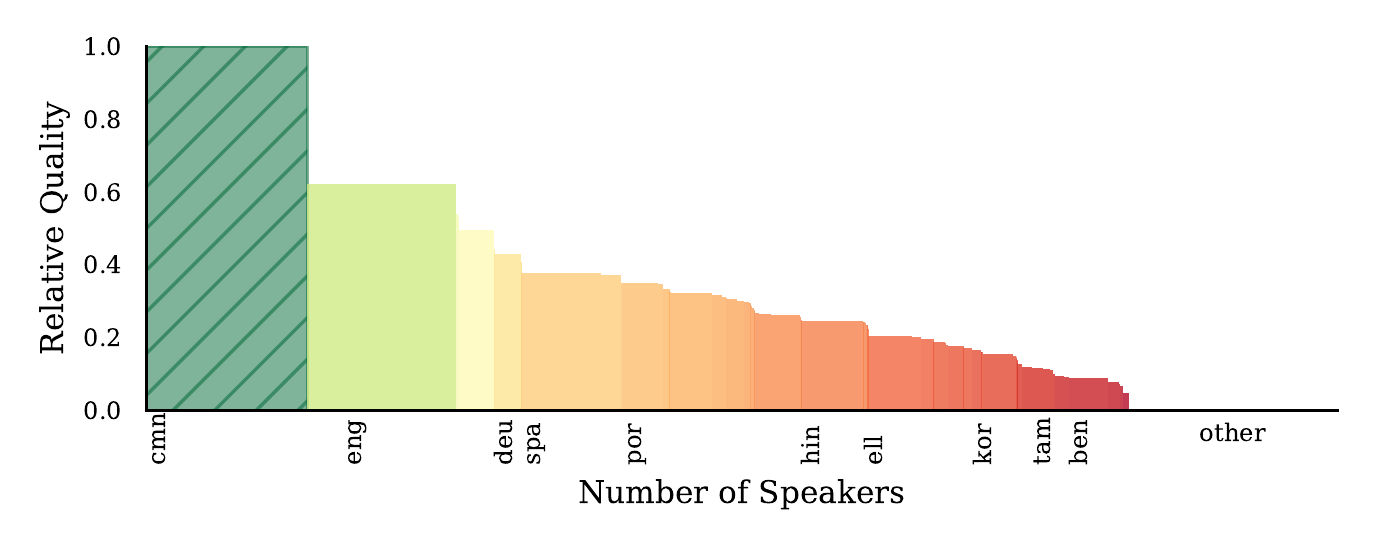} & \includegraphics[width=.32\linewidth]{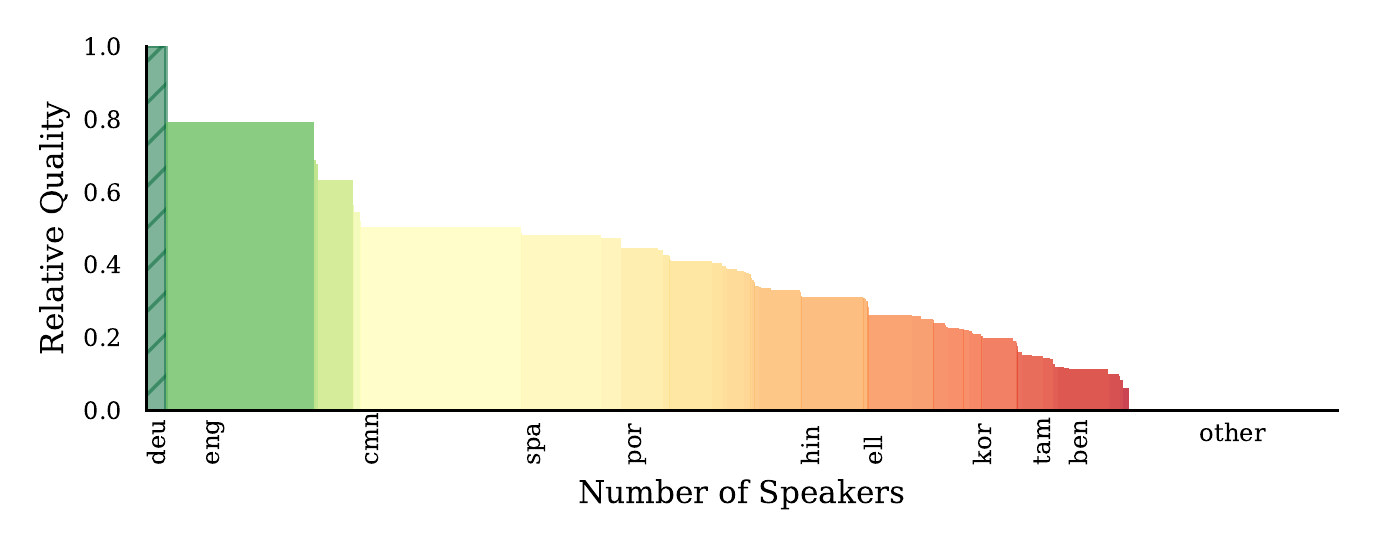} \\
    ell $\rightarrow  X$ & eng $\rightarrow  X$ & fin $\rightarrow  X$ \\
    \includegraphics[width=.32\linewidth]{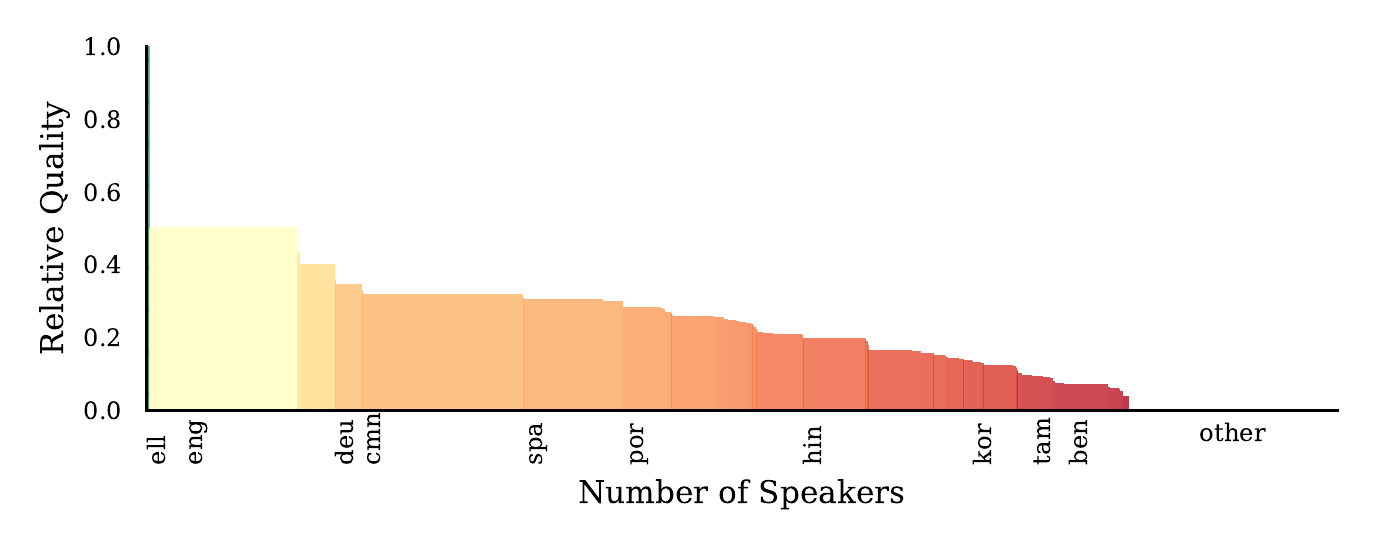} & \includegraphics[width=.32\linewidth]{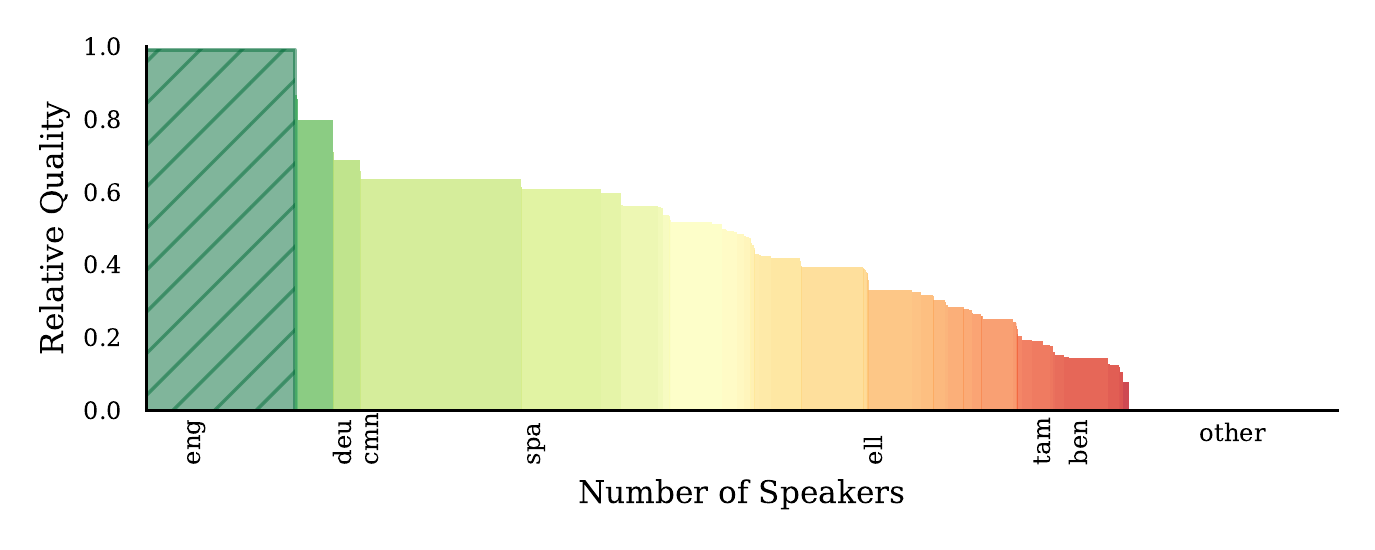} & \includegraphics[width=.32\linewidth]{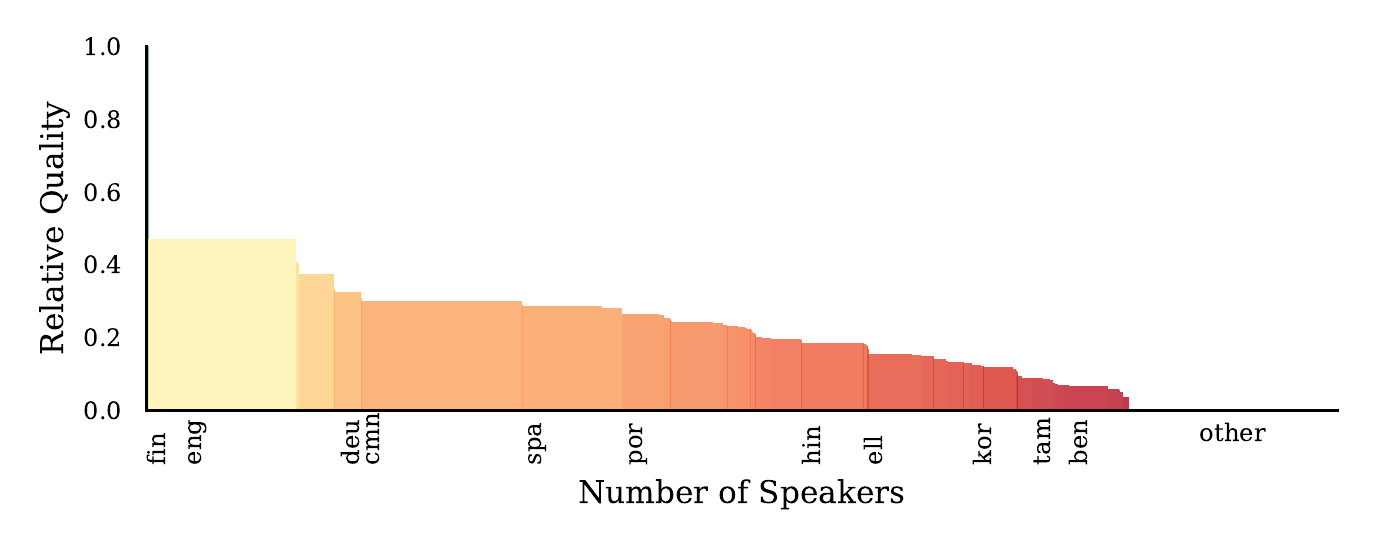} \\
    fra $\rightarrow  X$ & glg $\rightarrow  X$ & hau $\rightarrow  X$ \\
    \includegraphics[width=.32\linewidth]{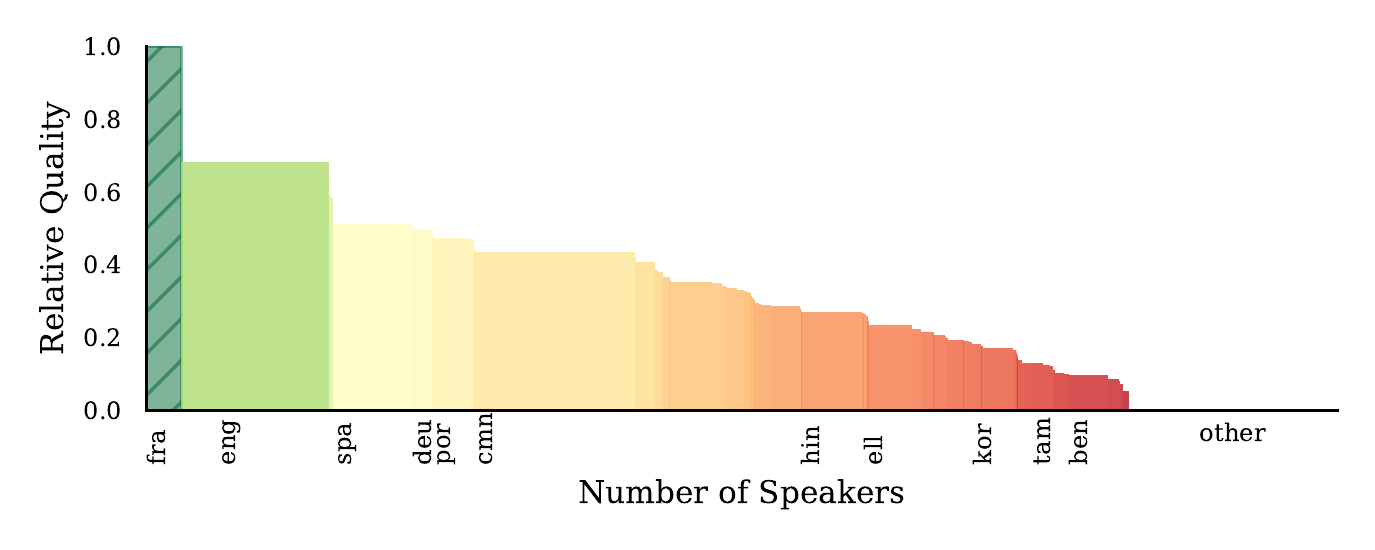} & \includegraphics[width=.32\linewidth]{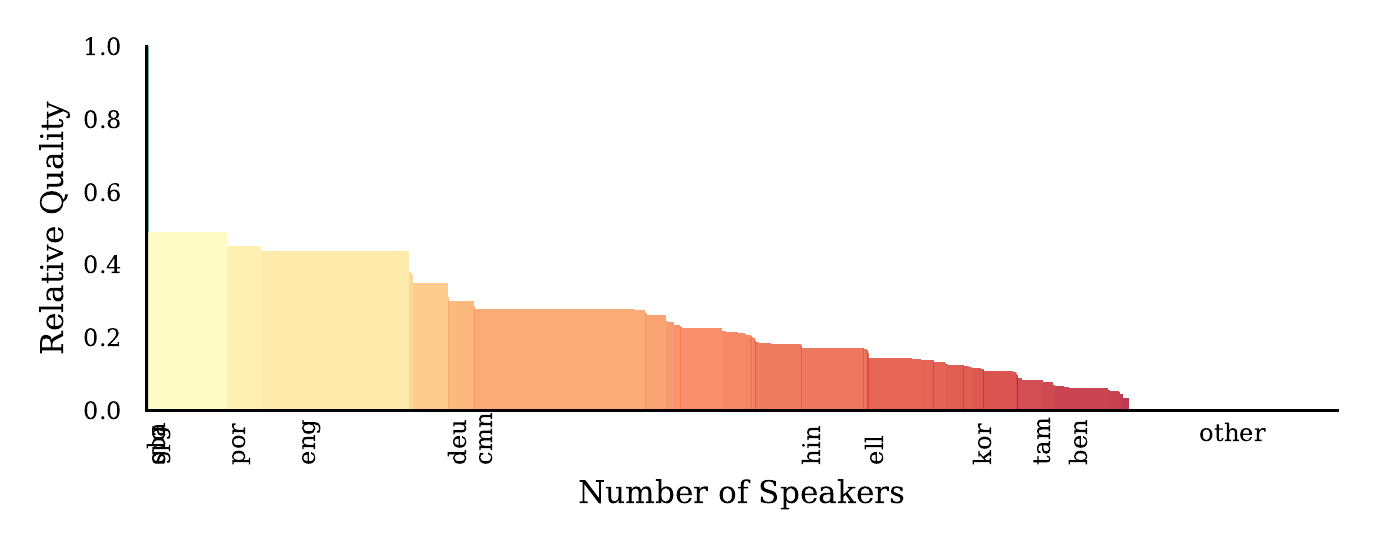} & \includegraphics[width=.32\linewidth]{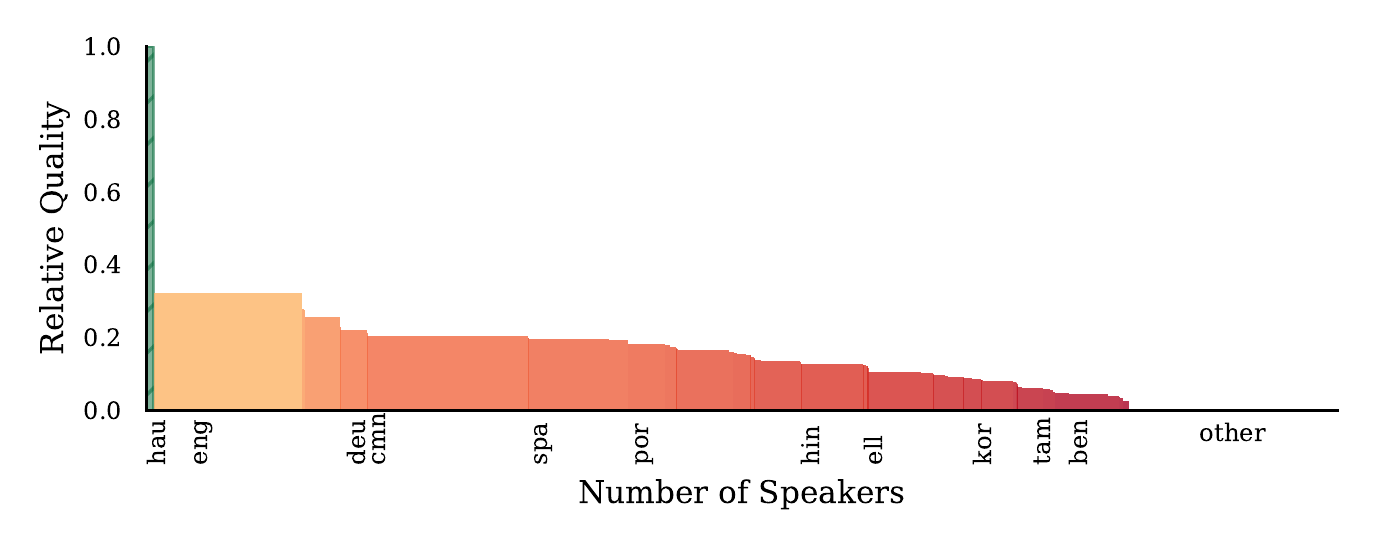} \\
    ita $\rightarrow  X$ & kin $\rightarrow  X$ & kor $\rightarrow  X$ \\
    \includegraphics[width=.32\linewidth]{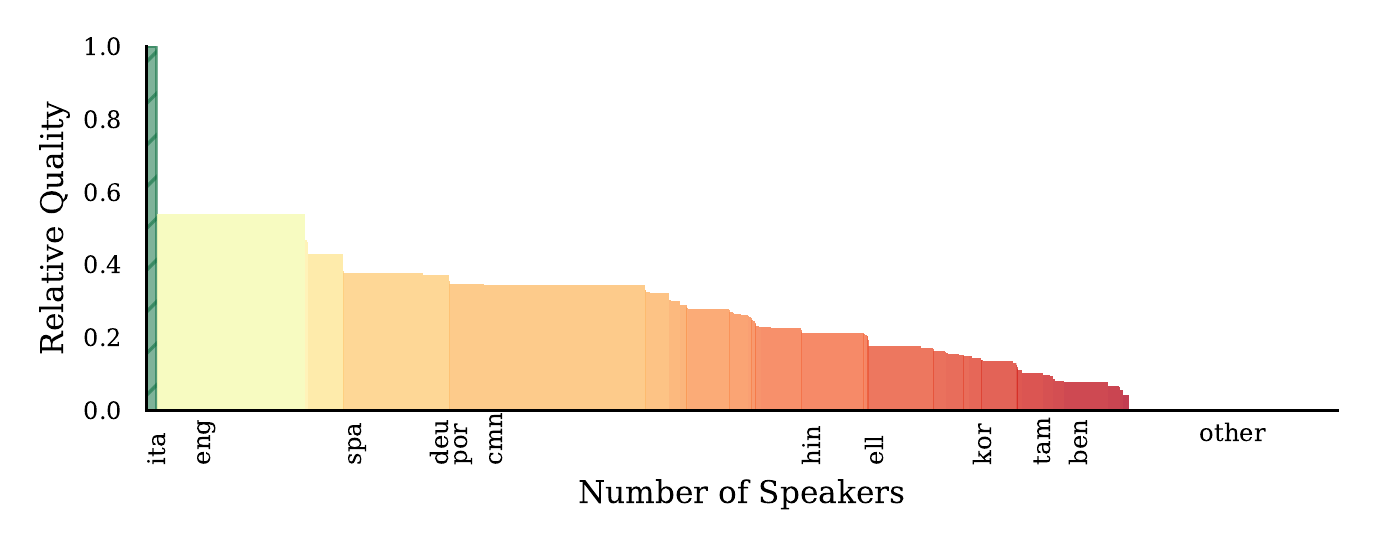} & \includegraphics[width=.32\linewidth]{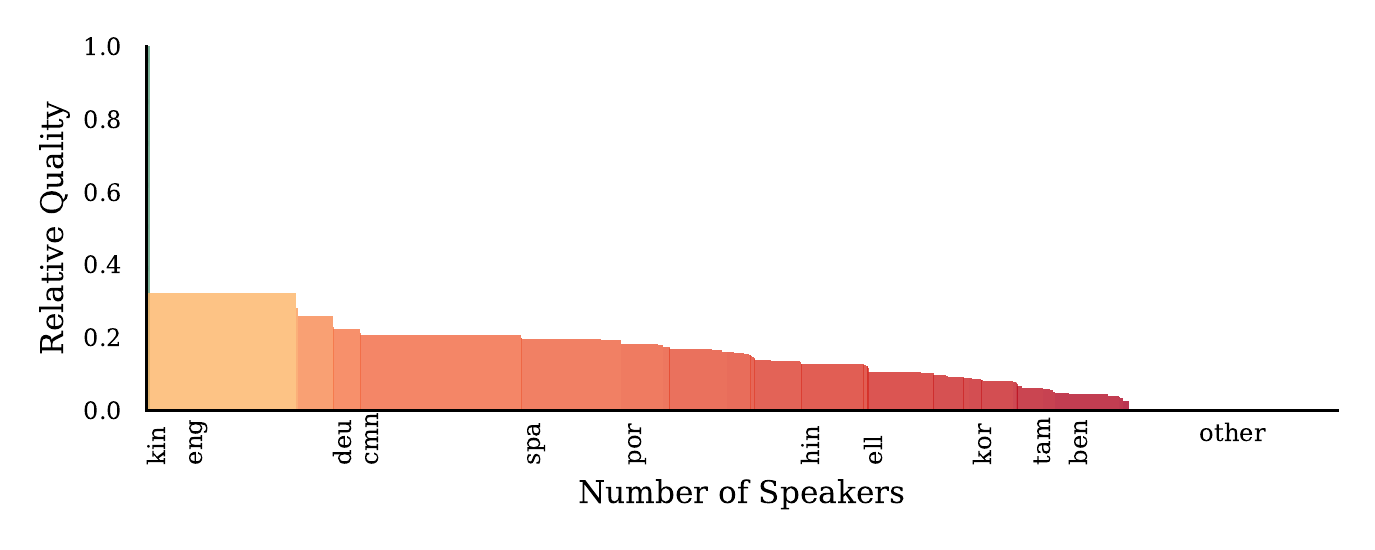} & \includegraphics[width=.32\linewidth]{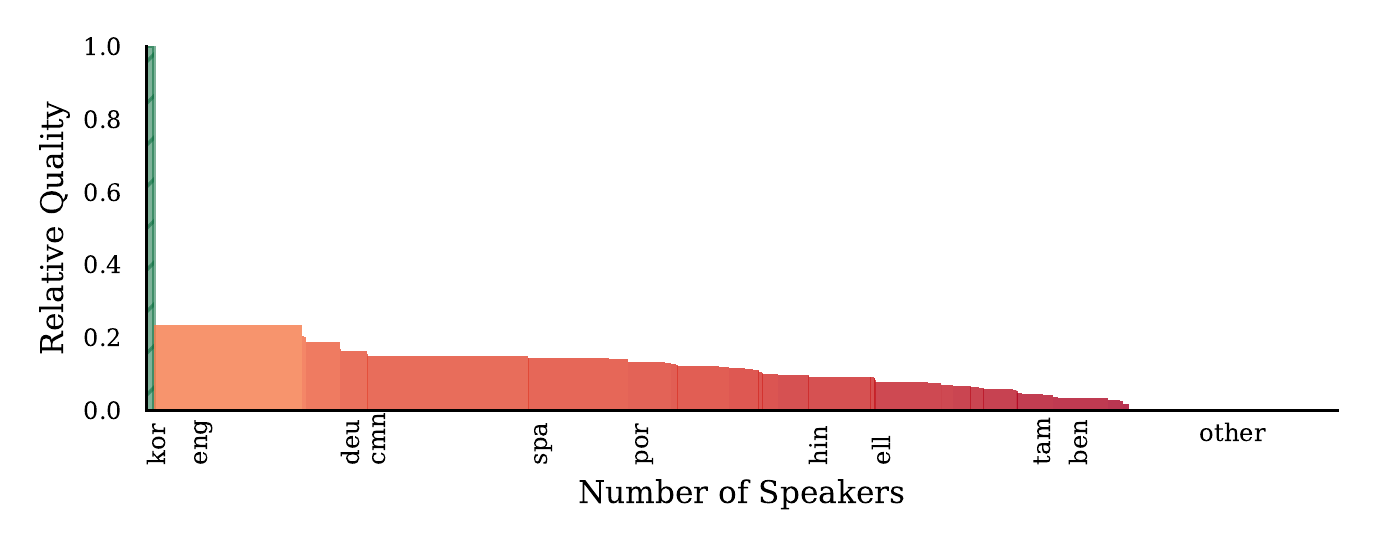} \\
    por $\rightarrow  X$ & rus $\rightarrow  X$ & spa $\rightarrow  X$ \\
    \includegraphics[width=.32\linewidth]{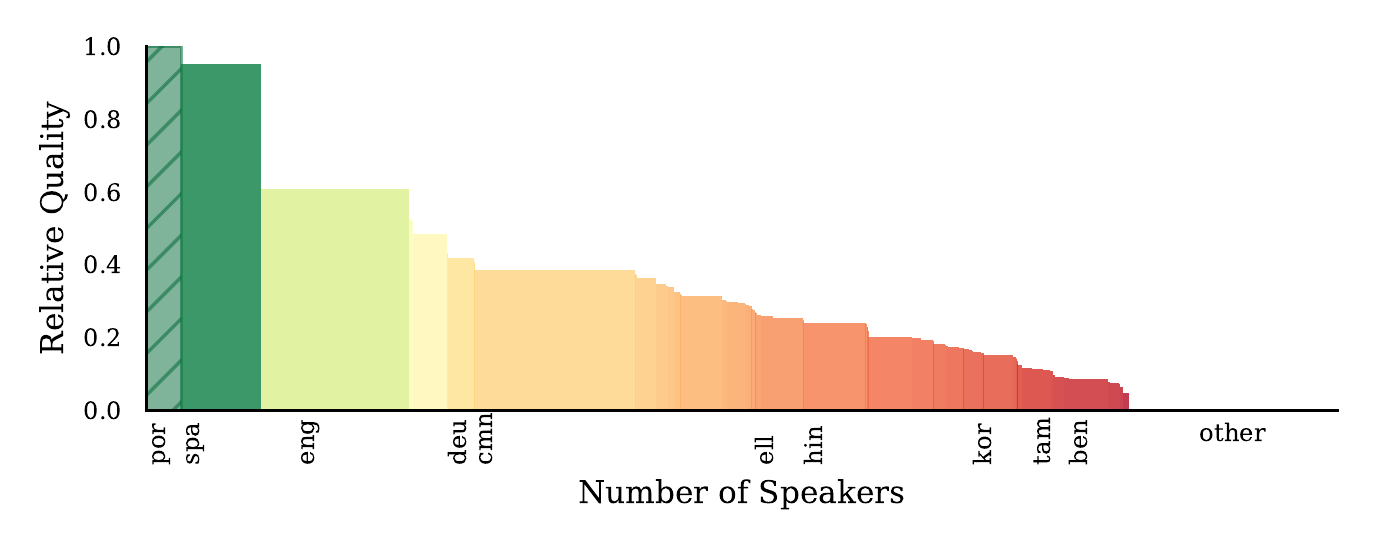} & \includegraphics[width=.32\linewidth]{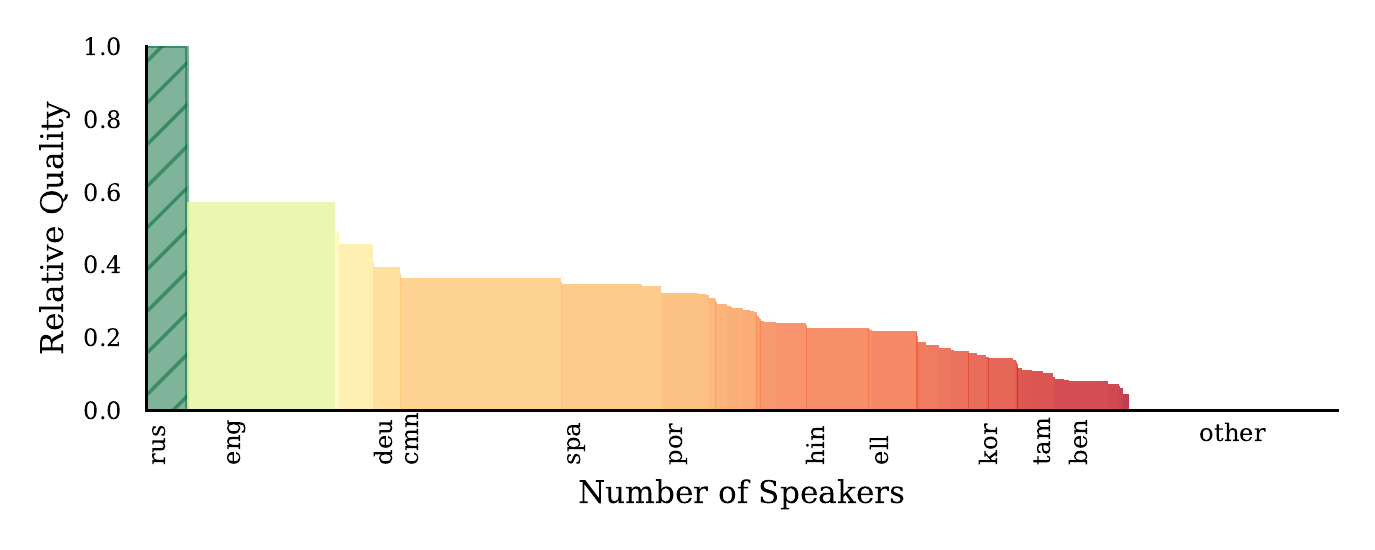} & \includegraphics[width=.32\linewidth]{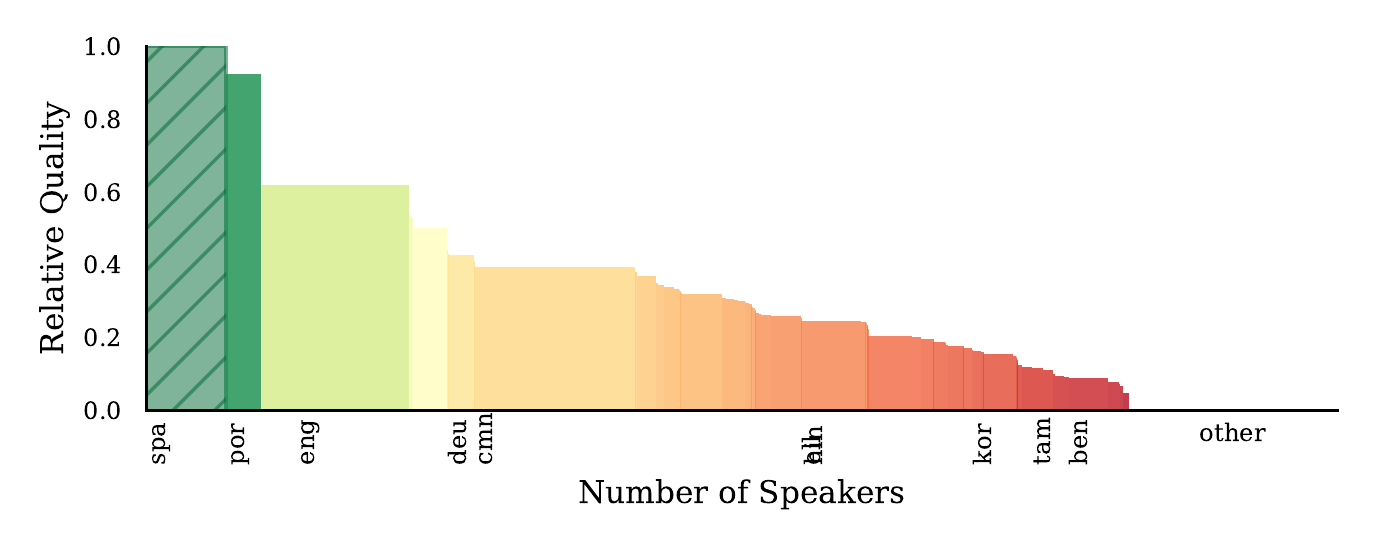} \\
    swa $\rightarrow  X$ & tam $\rightarrow  X$ & tur $\rightarrow  X$ \\
    \includegraphics[width=.32\linewidth]{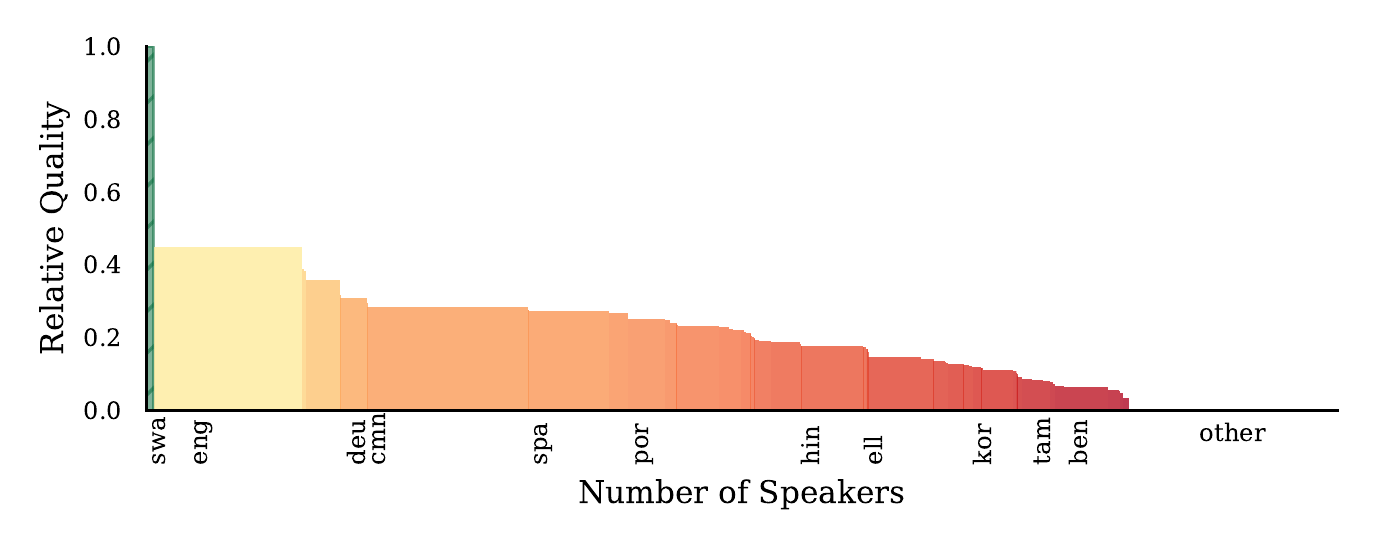} & \includegraphics[width=.32\linewidth]{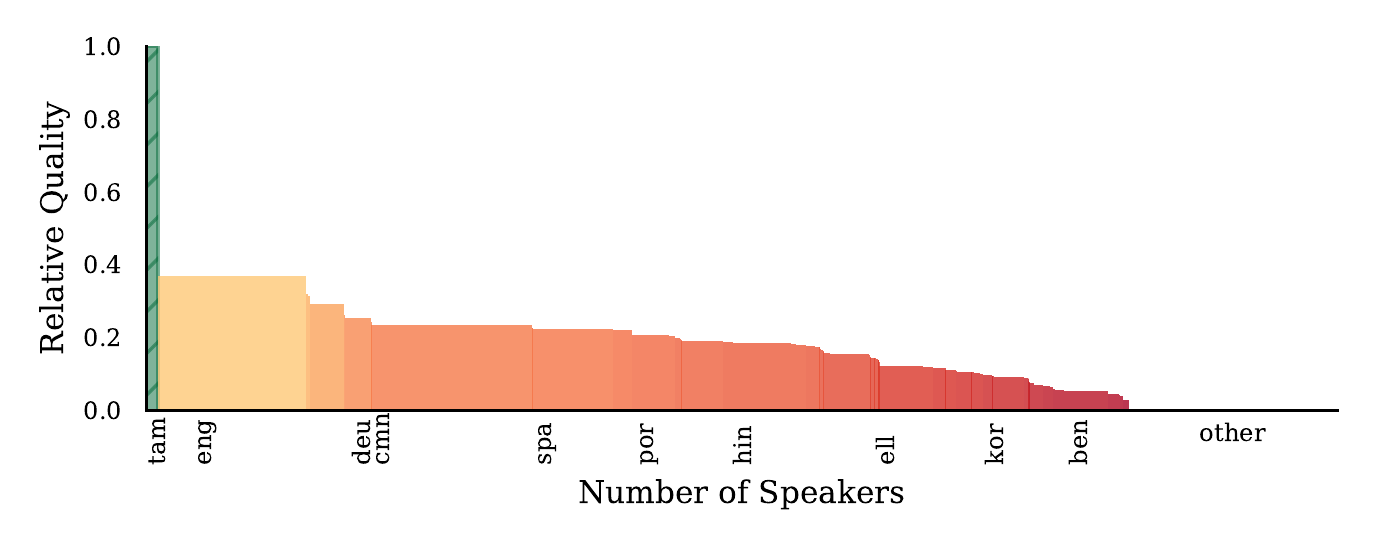} & \includegraphics[width=.32\linewidth]{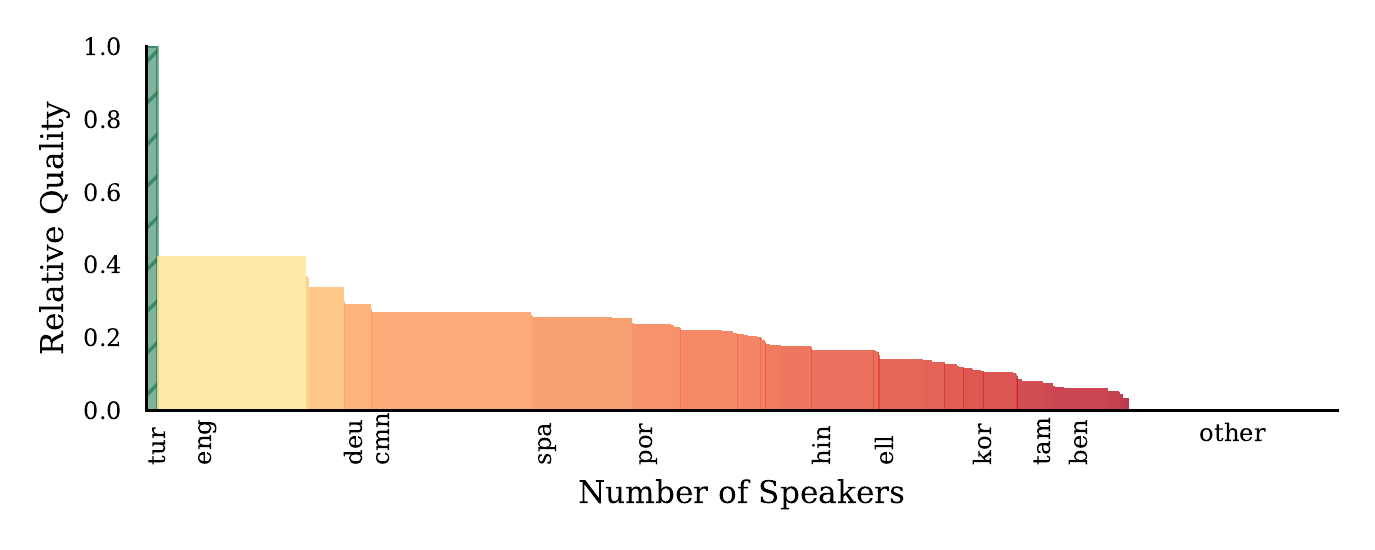} \\
    uig $\rightarrow  X$ & vie $\rightarrow  X$ & zul $\rightarrow  X$ \\
    \includegraphics[width=.32\linewidth]{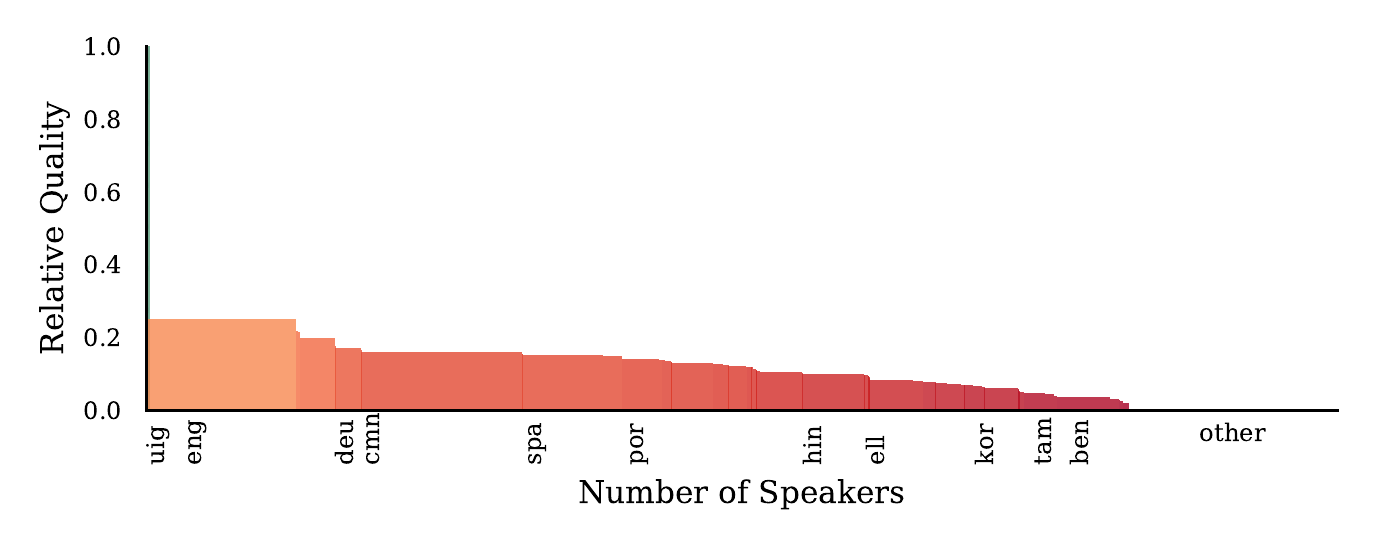} & \includegraphics[width=.32\linewidth]{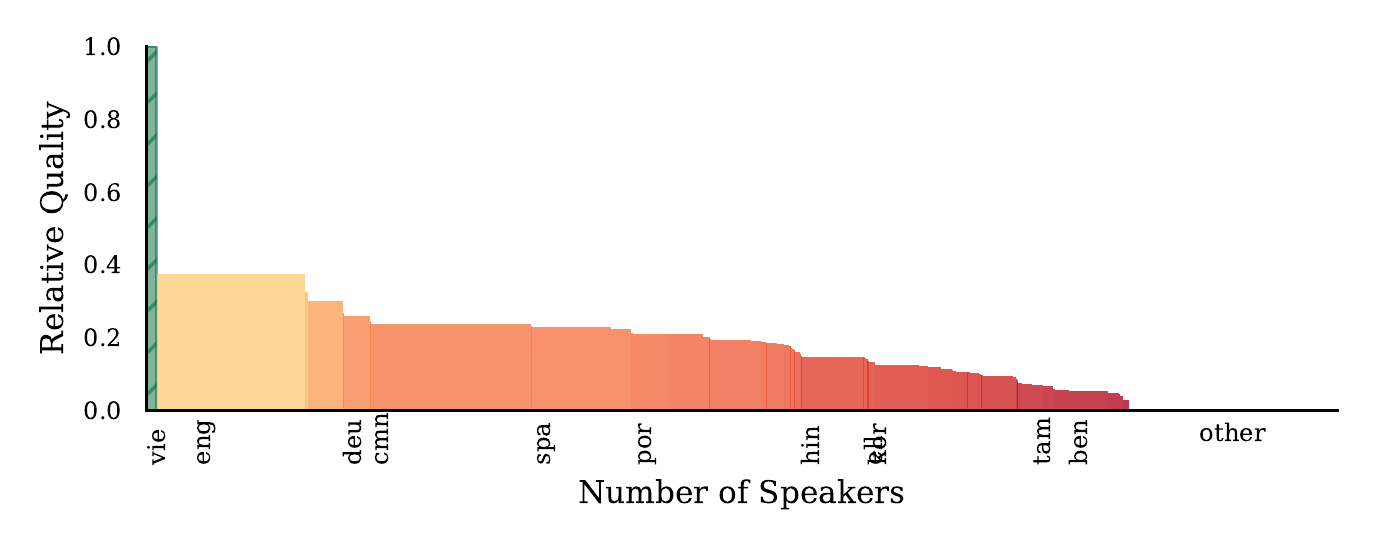} & \includegraphics[width=.32\linewidth]{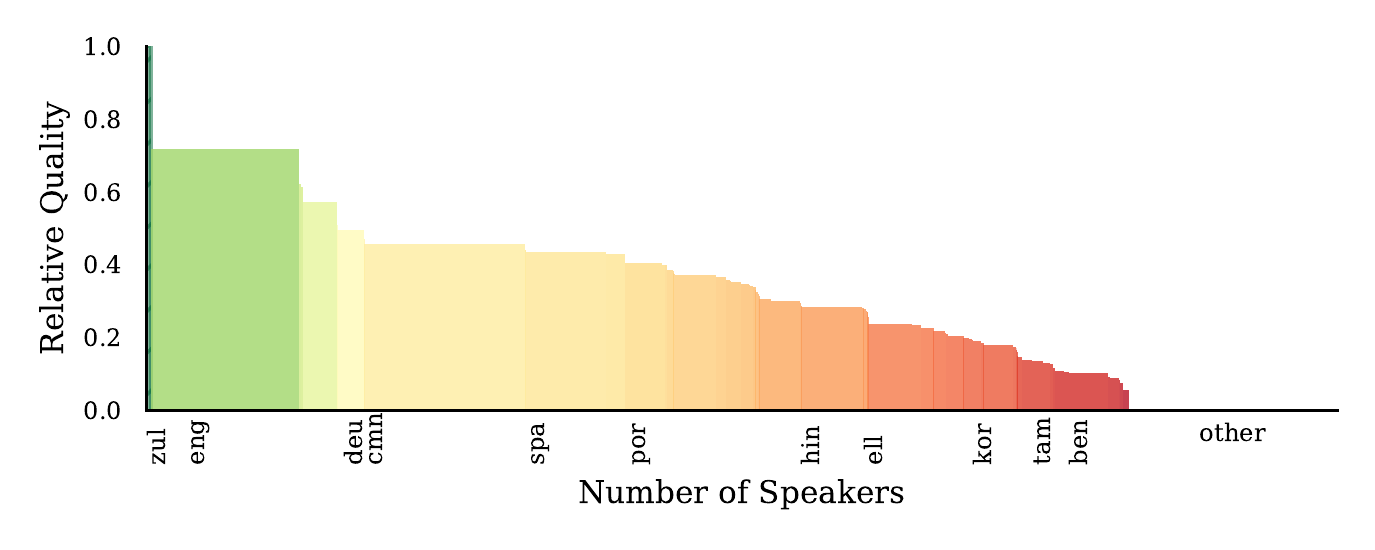} \\
    \end{tabular}
    \caption{Visualization of our measure on translation from 24 diverse languages.}
    \label{fig:mtfromlang}
\end{figure*}

Perhaps unexpectedly, the best (and often only) pivot is English in almost all cases. As a result, the final utility for a language X is very much dependent on the utility of the X-Eng (or Eng-X) systems. This is reflected by our scores for averaged by demographics and languages being very similar to the ones when we only focused on English. 
Nevertheless, the differences between scores for different languages are stark: the demographic-averaged utility for populous, well-studied languages like German ($M_1=0.356$), Chinese ($M_1=0.232$), or French ($M_1=0.309$) is almost double than under-served ones like Bengali ($M_1=0.148$), isiXhosa ($M_1=0.156$), Amharic ($M_1=148$), or Burmese ($M_1=0.092$). Figure~\ref{fig:mtfromlang} visualizes the different scores for translation from 24 languages under the demographic focus ($\tau=1$).

\begin{figure*}
    \centering
    \includegraphics[width=1\linewidth]{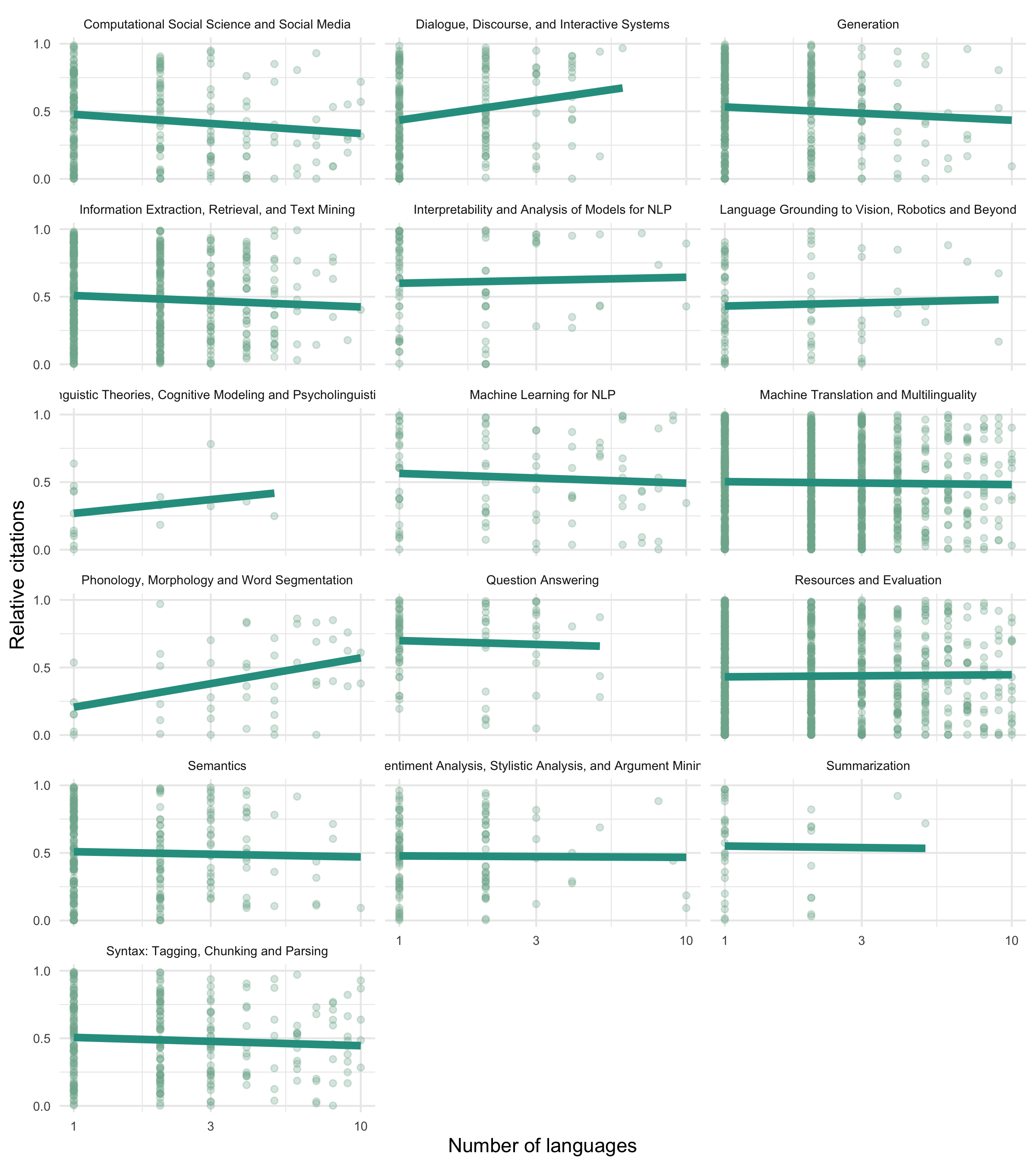}
        \caption{Cumulative citations vs number of languages in publications according to topic}
    \label{fig:fields_pubs}
\end{figure*}

\begin{table}[t]
    \centering
\begin{tabular}{ccccc}
    \toprule
    \multirow{2}{*}{rank} & \multirow{2}{*}{Lang.} & $\mathrm{pop}_{-\mathrm{eng}}$ & \multicolumn{1}{c}{Number of Studies} \\
     & & (M) & X--eng/eng--X \\
     \midrule
1 & cmn & 908.8 & 16/ 4 \\ 
2 & spa & 358.8 & 5/6 \\ 
3 & hin & 299.5 & 3/1 \\ 
4 & ben & 232.8 & 2/0 \\ 
5 & por & 207.7 & 3/3 \\ 
6 & ara & 205.4 & 9/6 \\ 
7 & rus & 145.6 & 9/6 \\ 
8 & jpn & 128.0 & 7/4 \\ 
9 & swa & 89.2 & 1/1 \\ 
10 & msa & 80.3 & 2/0 \\ 
11 & kor & 77.3 & 4/0 \\ 
12 & vie & 76.0 & 4/6 \\ 
13 & mar & 73.0 & 2/0 \\ 
14 & tam & 72.0 & 2/0 \\ 
15 & tur & 65.9 & 9/4 \\ 
16 & guj & 48.3 & 1/1 \\ 
17 & fra & 47.1 & 12/17 \\ 
18 & ind & 43.4 & 2/0 \\ 
19 & ita & 42.8 & 8/6 \\ 
20 & urd & 35.0 & 2/0 \\ 
21 & mya & 31.4 & 2/0 \\ 
22 & mal & 30.7 & 0/0 \\ 
23 & deu & 30.4 & 25/33 \\ 
24 & orm & 28.0 & 1/0 \\ 
25 & uzb & 27.9 & 0/0 \\ 
26 & ukr & 27.3 & 3/1 \\ 
27 & pol & 25.0 & 2/0 \\ 
28 & aze & 19.5 & 5/2 \\ 
29 & sin & 17.6 & 1/1 \\ 
30 & ron & 16.8 & 13/11 \\
\bottomrule
    \end{tabular}
    \caption{Machine Translation research interests on to and from English do not match our population-based demand model.}
    \label{tab:demand_pop}
\end{table}

\begin{table}[t]
    \centering
\begin{tabular}{cc|c}
    \toprule
    \multirow{2}{*}{Rank} & \multicolumn{2}{c}{Based on}\\
    & Imports & Exports\\
    \midrule
 1 & rus--bel & bel--rus \\ 
 2 & rus--kaz & mon--cmn \\ 
 3 & rus--hye & sqi--ita \\ 
 4 & rus--mon & hye--rus \\
 5 & rus--cmn & tgl--jpn \\ 
 6 & spa--som & nep--hin \\ 
 7 & hin--nep & aze--ita \\ 
 8 & ita--sqi & srp--bos \\ 
 9 & lit--lav &  lav--lit \\ 
10 & rus--aze & msa--jpn \\ 
11 & cmn--mya & lit--rus \\ 
12 & rus--fin & mya--cmn \\ 
13 & rus--ukr & est--fin \\ 
14 & cmn--tha & bos--hrv \\
15 & jpn--tgl & kat--rus \\
\bottomrule
    \end{tabular}
    \caption{Top-15 translation pairs based on demand estimated from economic indicators (import (export) partner share of the target (source) language).}
    \label{tab:demand_econ}
\end{table}

\begin{table}[t]
    \centering
\begin{tabular}{c|ccc}
    \toprule
        Language & \multicolumn{2}{c}{BLEU Score} & \multirow{2}{*}{Pivot} \\
        Pair & Estimated & Published & \\
    \midrule
slv--srp & 37.09 & 25.45 & eng--hrv \\ 
eng--nep & 10.56 & 6.8 & guj--hin \\ 
eng--hrv & 60.80 & 42.15 & srp \\ 
eng--hin & 13.78 & 12.5 & guj \\ 
hrv--eng & 50.42 & 48.07 & srp \\ 
ron--deu & 29.36 & 18.4 & eng \\ 
ron--fra & 33.98 & 26.53 & eng \\ 
ces--rus & 17.56 & 16.2 & eng \\ 
ces--deu & 23.36 & 19.3 & eng \\ 
ces--fra & 27.04 & 18.1 & eng \\ 
ita--deu & 26.08 & 19.85 & eng \\ 
rus--ces & 18.19 & 14.4 & eng \\ 
pol--ces & 9.90 & 7.2 & eng \\ 
nld--deu & 25.0 & 21.06 & eng \\ 
heb--fra & 27.41 & 23.25 & eng \\ 
srp--slv & 52.09 & 35.39 & hrv \\ 
deu--ron & 27.25 & 16.27 & eng \\ 
deu--ces & 25.19 & 20.1 & eng \\ 
deu--ita & 28.42 & 18.56 & eng \\ 
deu--nld & 26.48 & 20.31 & eng \\ 
deu--fra & 44.27 & 37.3 & eng \\ 
fra--ron & 23.52 & 19.3 & eng \\ 
fra--ces & 21.73 & 13.7 & eng \\ 
fra--heb & 18.88 & 13.54 & eng \\ 
spa--ces & 17.83 & 15.2 & por--eng \\ 
ara--fra & 26.83 & 25.07 & eng \\ 
slv--hrv & 55.64 & 40.44 & eng--srp \\
    \bottomrule
    \end{tabular}
    \caption{Translation pairs with a pivoting estimated utility (BLEU score) higher than the published result.}
    \label{tab:pivoting}
\end{table}

\begin{table}[t]
    \centering
\resizebox{\linewidth}{!}{
    \begin{tabular}{cc|cc}
    \toprule
        \multicolumn{2}{c}{Parameter} & ELDP difference & SE \\
        Negbinomial & Zero-inflated &  & \\
    \midrule
log-GDP & log-GDP & 0 & 0 \\ 
log-GDP & log-Users & -20.2 & 6.3 \\ 
log-Users & log-GDP & -31.9 & 9.8 \\ 
log-Users & log-Users & -69.8 & 13.2 \\ 
log-GDP & Fixed & -87.9 & 15.1 \\ 
log-Users & Fixed & -125.2 & 17.4 \\ 
Fixed & log-GDP & -263.3 & 40.7 \\ 
Fixed & log-Users & -307.9 & 41.9 \\ 
Fixed & Fixed & -437.1 & 46.9 \\ 
    \bottomrule
    \end{tabular}
    }
    \caption{ELDP model selection for GDP and number of user analysis, ordered from top (best) to bottom (worst).}
    \label{tab:gdp_pop}
\end{table}

\end{document}